\documentclass{article} 
\usepackage{iclr2025_conference,times}

\usepackage{amsmath,amsfonts,bm}

\def\eqref#1{equation~\ref{#1}}

\def\1{\bm{1}}

\DeclareMathAlphabet{\mathsfit}{\encodingdefault}{\sfdefault}{m}{sl}
\SetMathAlphabet{\mathsfit}{bold}{\encodingdefault}{\sfdefault}{bx}{n}

\usepackage{hyperref}
\usepackage{url}
\usepackage{amssymb}
\usepackage{amsmath} 
\usepackage{tcolorbox}
\usepackage{adjustbox}
\usepackage{booktabs}
\usepackage{makecell}
\usepackage{array}
\usepackage{subcaption,caption}
\usepackage{pifont}

\newtheorem{lemma}{Lemma}

\usepackage{enumitem} 

\usepackage{wrapfig}

\usepackage{multirow}

\title{COIN: Uncertainty-Guarding Selective Question Answering for Foundation Models with Provable Risk Guarantees}

\author{
 \textbf{Zhiyuan Wang}\textsuperscript{1}, 
 \textbf{Jinhao Duan}\textsuperscript{2}, 
 \textbf{Qingni Wang}\textsuperscript{1}, 
 \textbf{Xiaofeng Zhu}\textsuperscript{1}, 
 \textbf{Tianlong Chen}\textsuperscript{3},\\
 \textbf{ Xiaoshuang Shi\textsuperscript{1$\ast$}},
 \textbf{Kaidi Xu\textsuperscript{2}\thanks{Corresponding Authors}}
\\
\\
 \textsuperscript{1}University of Electronic Science and Technology of China
\\
 \textsuperscript{2}Drexel University
\\
 \textsuperscript{3}University of North Carolina at Chapel Hill
\\
\small \texttt{\{yhzywang, qingni1031, seanzhuxf, xsshi2013\}@gmail.com} \\
\small \texttt{\{jd3734, kx46\}@drexel.edu}
\quad \texttt{tianlong@cs.unc.edu}
}

\iclrfinalcopy 
\begin{document}

\maketitle

\begin{abstract}
Uncertainty quantification (UQ) for foundation models is essential to identify and mitigate potential hallucinations in automatically generated text. 
However, heuristic UQ approaches lack formal guarantees for key metrics such as the false discovery rate (FDR) in selective prediction.
Previous work adopts the split conformal prediction (SCP) framework to ensure desired coverage of admissible answers by constructing prediction sets, but these sets often contain incorrect candidates, limiting their practical utility. 
To address this, we propose COIN, an uncertainty-guarding selection framework that calibrates statistically valid thresholds to filter a single generated answer per question under user-specified FDR constraints.
COIN estimates the empirical error rate on a calibration set and applies confidence interval methods such as Clopper–Pearson to establish a high-probability upper bound on the true error rate (i.e., FDR). 
This enables the selection of the largest uncertainty threshold that ensures FDR control on test data while significantly increasing sample retention. 
We demonstrate COIN's robustness in risk control, strong test-time power in retaining admissible answers, and predictive efficiency under limited calibration data across both general and multimodal text generation tasks. 
Furthermore, we show that employing alternative upper bound constructions and UQ strategies can further boost COIN's power performance, which underscores its extensibility and adaptability to diverse application scenarios. 
\end{abstract}

\section{Introduction}
\label{sec: Introduction}
Recent advances in foundation models, including large language models (LLMs) and large vision-language models (LVLMs), have remarkably propelled progress in various downstream tasks, especially question answering (QA)~\cite{hurst2024gpt,zhao2025can,guo2025deepseek}. 
However, such models remain susceptible to trustworthiness issues like hallucination, prone to responding with plausible yet incorrect information~\cite{huang2025survey,duan2025truthprint}. 
Such deficiencies undermine their reliable deployment in risk-sensitive domains~\cite{yao2024survey,zheng2025large,penny2025reducing}. 
A prevalent mitigation strategy is to estimate model uncertainty during communication, thereby enabling selective rejection or abstention when predictions exhibit high uncertainty~\cite{lingenerating,NEURIPS2024_10c456d2,wang2025word,hou-etal-2025-probabilistic}. 

Despite their empirical effectiveness, heuristic uncertainty quantification (UQ) methods lack formal guarantees for controlling critical metrics such as the false discovery rate (FDR) in selective generation tasks~\cite{chen-etal-2023-adaptation,tayebati2025learning}. 
Recent studies have successfully adapted split conformal prediction (SCP)~\cite{angelopoulos2021gentle}, which converts heuristic uncertainty estimates from arbitrary models to statistically rigorous ones by constructing prediction sets, to QA tasks, ensuring coverage of admissible responses at a user-specified risk level~\cite{wang-etal-2024-conu,ye2024benchmarking,quach2024conformal,wang2025sample}. 
However, such prediction sets typically include unreliable candidates, severely limiting their practical effectiveness~\cite{Jesse2025Disparate}. 
Consequently, determining statistically valid uncertainty thresholds that rigorously control FDR for selective prediction tasks requiring a single definitive answer per input remains an open challenge. 

\begin{figure}[!t]
    \centering
    \includegraphics[width=\linewidth]{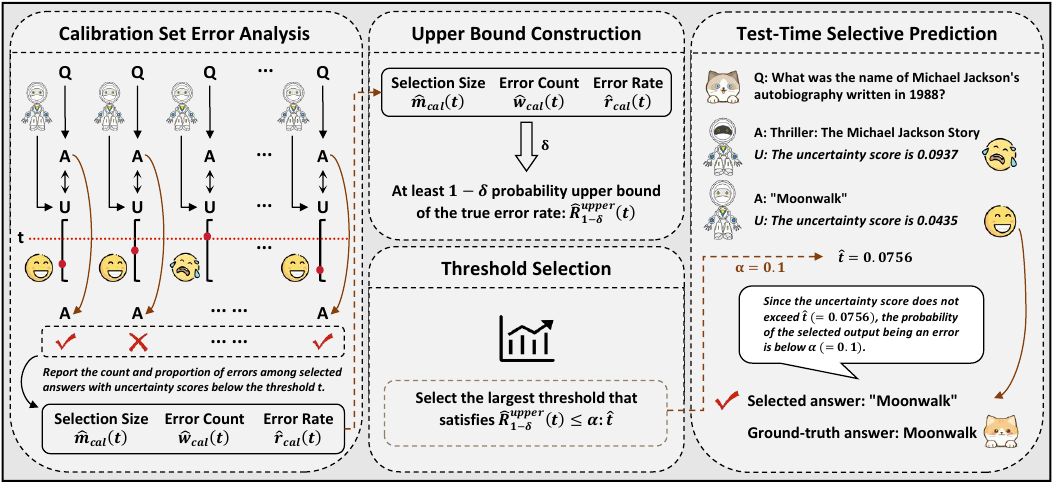}
    \caption{An overview of the three-stage COIN framework and an illustration of the test-time selective prediction with the guarantee of error rate.}
    \label{fig: Overview}
\end{figure}

A previous work develops conformal alignment (CA)~\cite{gui2024conformal}, which first trains an alignment predictor and then applies the conformal p-value framework to filter new test samples~\cite{jin2023selection,huang2024confine}. 
An answer is selected only when its associated predicted alignment score exceeds a data-dependent threshold. 
This conformalized selection mechanism ensures FDR control among the selected QA samples. 
However, CA requires traversing the entire calibration set to compute the p-value for each test sample, and depends on the Benjamini–Hochberg (BH) procedure~\cite{benjamini1995controlling} to determine the p-value threshold, making it highly time-consuming for large-scale QA tasks and causing unnecessary rejection of many correct answers. 

In this paper, we propose \textbf{COIN}, a three-stage framework for uncertainty-guarding selective prediction, comprising \textbf{C}alibration set error analysis, upper b\textbf{O}und construct\textbf{I}on, and threshold selectio\textbf{N}, as shown in Figure~\ref{fig: Overview}. 
Each stage is implemented as a modular component to facilitate separate optimization and flexible substitution. 
Following the principles of SCP-based frameworks, COIN begins by holding out a dedicated calibration set for threshold calibration. 
Given a candidate threshold, we quantify the uncertainty for each calibration data, retaining only those samples whose uncertainty scores fall below the threshold, i.e., treating them as trustworthy. 
Subsequently, we estimate the empirical error rate over this uncertainty-filtered calibration subset. 
To rigorously constrain selection risks based on test statistics from the in-distribution calibration data, COIN proceeds to construct a high-probability upper confidence bound of the true failure/error rate, computed from observations of the selected subset. 
Finally, we identify the largest threshold whose corresponding upper bound remains below the risk level, thereby providing robust control over the FDR on test data while maximizing sample retention. 

Specifically, in the first stage, we record the size of the uncertainty-filtered calibration subset and the number of failure cases (inadmissible answers), from which the empirical failure rate is computed. 
These statistics serve as the foundation for the second stage, where we apply the one-sided Clopper–Pearson interval~\cite{clopper1934use} to derive an exact, high-probability upper bound on the true failure rate. 
Owing to COIN’s modular and flexible design, alternative concentration inequalities such as Hoeffding’s inequality~\cite{hoeffding1994probability,bentkus2004hoeffding,bates2021distribution} can also be employed to construct upper confidence bounds. 
While these alternatives are not exact, they significantly improve computational efficiency in large-scale QA scenarios while maintaining valid statistical guarantees. 
With the high-probability upper bound in place, the third stage of COIN focuses on threshold selection. 
By progressively increasing the uncertainty threshold, we aim to determine the maximum value such that the associated upper bound on the failure rate does not exceed a user-specified risk level. 
This choice ensures rigorous control over test-time risk while maximizing the number of selected admissible answers, striking a trade-off between rigor and efficiency. 

COIN applies to diverse text generation tasks (\textbf{generality}). 
In this study, we evaluate its statistical validity on two textual QA datasets: closed-ended CommonsenseQA~\cite{talmor-etal-2019-commonsenseqa} and open-domain TriviaQA~\cite{joshi-etal-2017-triviaqa}, employing five LLMs. 
Additionally, we also implement COIN on the open-domain multimodal MMVet dataset~\cite{yu2024mmvet} across five LVLMs. 
Empirical results confirm that COIN significantly outperforms CA in retention of admissible answers (\textbf{power}) while constraining the FDR at desired risk levels (\textbf{rigor}). 
Moreover, owing to COIN's \textbf{customizable} structure, we employ different UQ methods under both white-box and black-box settings in the first stage, to obtain the most accurate filtering, and select the optimal upper bound construction strategy in the second stage, to balance computational load and sample retention. 
Furthermore, we validate COIN's \textbf{robustness} and predictive \textbf{efficiency} across varying splits of calibration and test data. 

The main contributions of this paper are summarized as follows:
\begin{itemize}[leftmargin=10pt]
    \item We investigate uncertainty-guarding selective prediction in general QA tasks with the constraint of FDR while maximizing sample retention, which is a previously under-explored topic. 
    \item COIN precomputes the maximal uncertainty threshold on the calibration set for a target risk level, substantially improving test-time selection efficiency. 
    \item Extensive evaluations on three QA datasets with different response formats and modalities demonstrate that COIN rigorously controls the failure rate of selected samples under the condition that uncertainty falls below the calibrated threshold while covering more correct answers. 
\end{itemize}

\section{Related Work}
\noindent\textbf{SCP in QA tasks.} 
Recent research has effectively applied the SCP framework to language models for reliable QA~\cite{campos-etal-2024-conformal}. 
Two influential studies~\cite{kumar2023conformal,ye2024benchmarking} adapt SCP to closed-ended QA tasks, providing statistical coverage guarantees while employing the size of the prediction set as a proxy for uncertainty. 
In open-domain settings, several studies~\cite{quach2024conformal,wang-etal-2024-conu,kaur2024addressing,wang2025sample} extend SCP by introducing conformalized uncertainty criteria that construct prediction sets with guaranteed coverage of admissible responses. 
While promising, both lines of research suffer from selection bias within prediction sets~\cite{jin2023selection,jin2024confidence}. 
For tasks that require a single answer per question, two works~\cite{mohri2024language,cherian2024large} attempt to ensure conformal factuality by filtering out unreliable sub-claims from model outputs, but such modification can remove many valuable and accurate claims, resulting in vague or uninformative responses. 

\noindent\textbf{Abstention.} 
Our problem setting falls under selective prediction, where the model is allowed to either generate an output or abstain when uncertain~\cite{kadavath2022language}. 
Prior studies have explored UQ techniques during communication to inform users of model output reliability~\cite{su-etal-2024-unsupervised,wang2025word,yang2025uncle,abbasli2025comparing}. 
For example, VL-Uncertainty~\cite{zhang2024vl} perturbs both the input images and questions to estimate the semantic entropy~\cite{kuhn2023semantic}. 
If the uncertainty score exceeds a predefined threshold, the model abstains, treating the output as a potential hallucination. 
BTPROP~\cite{hou-etal-2025-probabilistic} constructs a belief tree of related claims and applies probabilistic reasoning to detect contradictions, thereby identifying possible hallucinations in the model's output. 
However, these abstention strategies lack formal statistical guarantees~\cite{yadkori2024mitigating,gui2024conformal}, such as controlling the FDR over selected samples. 

To address this, CA~\cite{gui2024conformal} fits an alignment function and applies conformalized selection to new samples. 
While CA controls FDR, it requires comparing each test sample against the entire calibration set to compute p-values, and its use of the BH procedure rejects many correct answers. 
Instead, we follow the idea of learn then test (LTT)~\cite{angelopoulos2021learn,quach2024conformal}, which calibrates a candidate set of parameters on the calibration data, each offering guaranteed risk control.  
This enables us to select the optimal parameter (threshold) that maximizes the selection of QA samples while maintaining the desired risk level. 
Furthermore, each new sample can be directly evaluated against a fixed threshold determined by the desired risk level, significantly improving inference efficiency. 
Building on this, we further integrate heuristic UQ methods and upper confidence bound constructions to develop a novel formulation that enables conditional risk guarantees. 
\section{Methodology}
\subsection{Preliminaries}\label{sec: preliminaries}
\noindent\textbf{Problem Formulation.} 
Let $F: \mathcal{X} \rightarrow \mathcal{Y}$ be an LLM, and let $U$ be a scalar function that estimates the model's uncertainty score for a given input. 
For each test question $X_{test} \in \mathcal{X}$ with its ground-truth answer $Y_{test}^* \in \mathcal{Y}$, the model generates a candidate answer $\hat{Y}_{test} \sim \mathbb{P}\left(Y \mid X_{test}\right)$, and we compute the associated uncertainty score as $U\left( F\left(X_{test}\right)\right)$. 
We consider an uncertainty-guarding deployment policy, where a candidate answer is retained only if the estimated uncertainty score falls below a specified threshold $t$, i.e., $U\left( F\left(X_{test}\right)\right) \leq t$. 
\textbf{Our goal} is to calibrate the threshold $t$ in a statistically rigorous way so that the probability of accepting an incorrect answer, conditioned on a low uncertainty score (i.e., $\leq t$), remains below a desired level with high probability. 
Formally, for a user-specified risk level $\alpha \in \left(0,1\right)$ and the confidence level $1-\delta$ (e.g., 0.99), we aim to guarantee  
\begin{equation}\label{eq: objective}
    \mathrm{Pr}\left( \mathrm{Pr}\left( A\left( \hat{Y}_{test}, Y_{test}^* \right) = 0 \mid U\left( F\left(X_{test}\right)\right) \leq t \right) \leq \alpha \right) \geq 1 - \delta,
\end{equation}
where $A\left( \hat{Y}, Y^* \right) \in \left\{0,1\right\}$ is an admission indicator function that returns $1$ if the candidate answer $\hat{Y}$ matches the corresponding ground truth $Y^*$ and $0$ otherwise. 
The inner probability measures the true conditional failure rate (TCFR) under the event $U\left( F\left(X\right)\right) \leq t$. 

To formalize this guarantee over random test instances, let $\left( X, Y^* \right)$ be drawn from a data-generating distribution $\mathcal{D}$, where $X\in \mathcal{X}$ is a random question variable and $Y^*\in \mathcal{Y}$ is the corresponding random ground-truth variable, and $\hat{Y}$ denote the model's output to $X$. 
We define a binary correctness variable $Z := \mathbf{1} \left\{ A\left( \hat{Y}, Y^* \right) = 1 \right\}$ indicating whether the candidate answer $\hat{Y}$ of $X$ is admissible ($Z=1$) or not ($Z=0$). 
With this notation, the TCFR under the threshold $t$ is defined as
\begin{equation}
    R\left(t\right) := \mathbb{E}\left[ 1-Z \mid U\left( F\left(X \right) \right) \leq t \right],
\end{equation}
and the statistical guarantee becomes 
\begin{equation}
    \mathrm{Pr}\left( R\left(t\right) \leq \alpha \right) \geq 1-\delta, 
\end{equation}
following a PAC-style risk control~\cite{snell2023quantile,zollo2024prompt,ni2025towards}, which forms the statistical foundation of our threshold calibration framework. 

Building on previous research on selective prediction~\cite{gui2024conformal,yadkori2024mitigating,wang2025sconu} and split conformal methods~\cite{wang-etal-2024-conu,quach2024conformal,shahrokhi2025conformal}, where data points $\left(X_i, Y_i^*\right)$ are assumed to be \textit{independent and identically distributed} (i.i.d.) samples drawn from the underlying data-generating distribution $\mathcal{D}$~\cite{angelopoulos2021learn,angelopoulos2021gentle,angelopoulos2024theoretical}, we next formalize the statistical structure induced by the i.i.d. data-generating process together with the uncertainty-guarding selection mechanism, which underlies the further estimation of the TCFR.

\noindent\textbf{Bernoulli Structure under Conditional Selection.} 
For a given uncertainty threshold $t$, we define a (random) subset of selected samples as $\mathcal{S}_{t} := \left\{ \left(X_i, Y_i^*\right): U\left( F\left(X_i \right) \right) \leq t \right\}$, which corresponds to a subpopulation drawn from the \textit{conditional distribution} $\mathcal{D}_t := \mathcal{D} \mid U(F(X)) \leq t$. 
In practice, we can compute the (random) empirical failure rate on this subset as 
\begin{equation}
    \hat{R}\left(t\right) = \frac{1}{\left| \mathcal{S}_{t} \right|} \textstyle\sum_{\left(X_i, Y_i^*\right) \in \mathcal{S}_{t}} \left(1-Z_i\right).
\end{equation} 
Considering that the TCFR $R(t)$ is unknown and cannot be directly evaluated on new test samples, we rely on its empirical counterpart $\hat{R}(t)$ observed from the selected subset $\mathcal{S}_t$ to construct a valid upper bound of $R(t)$ that holds with high probability. 
To enable statistically sound estimation, we follow the standard assumption that \textit{the original QA data points $\left(X_i, Y_i^*\right)$ are i.i.d. samples from the data-generating distribution $\mathcal{D}$}, which is widely adopted in recent risk control frameworks of QA tasks~\cite{kumar2023conformal,wang-etal-2024-conu,ye2024benchmarking,gui2024conformal,wang2025sconu,ni2025towards}. 
We then introduce the following lemma. 

\begin{lemma}[Bernoulli Property of Correctness Indicators]\label{lm: i.i.d. brv}
Let $\left(X_i, Y_i^*\right)$ be i.i.d. samples drawn from a joint distribution $\mathcal{D}$, where $Y_i^* \sim \mathbb{P}\left(Y \mid X_i\right)$. 
Suppose that the uncertainty-guarding selection indicator is defined as $I_i := \mathbf{1}\left\{U(F(X_i)) \le t\right\}$, where $U(F(X_i))$ is a deterministic function of $X_i$, and the correctness indicator is $Z_i := \mathbf{1} \left\{ A\left( \hat{Y}_i, Y_i^* \right) = 1 \right\}$, where $\hat{Y}_i\sim \mathbb{P}\left(Y \mid X_i\right)$, then the subset $\mathcal{S}_{t}$ corresponds to i.i.d. samples from the conditional distribution $\mathcal{D}_t$, and the correctness indicators $\left\{Z_i\right\}_{\left(X_i, Y_i^*\right)\in \mathcal{S}_{t}}$ are i.i.d. Bernoulli random variables with success probability
\begin{equation}
         p_t:= 1-R\left(t\right),
\end{equation}
where $R\left(t\right)$ is the TCFR. 
Equivalently, the failure indicators $W_i := 1 - Z_i \sim \mathrm{Bernoulli} \left(R\left(t\right) \right)$ are also i.i.d. Bernoulli random variables under this conditional distribution $\mathcal{D}_t$. 
\end{lemma}

This statistical structure is directly implied by the i.i.d. data-generating process and the uncertainty-guarding selection, without introducing additional assumptions~\cite{angelopoulos2021learn,quach2024conformal,ni2025towards,shahrokhi2025conformal}. 
A formal justification is provided in Appendix~\ref{appendix: statistical Structure}. 
We next employ the empirical failure rate $\hat{R}(t)$ as an unbiased and consistent estimator of the TCFR $R(t)$, enabling the construction of a statistical upper bound with high probability.

\subsection{Risk-Bounded Threshold Calibration}
Building on the statistical structure established above, we now present the full calibration procedure of COIN for selecting a statistically valid uncertainty threshold, denoted as $\hat{t}$, that ensures a high-confidence guarantee on the TCFR $R\left(t\right)$. 
Our approach follows the standard split conformal frameworks by reserving a held-out set of $M$ i.i.d. calibration data points, $\mathcal{D}_{cal} = \left\{ \left( x_i, y_i^* \right) \right\}_{i=1}^{M} \subset  \mathcal{D}$, 
where $x_i \in \mathcal{X}$ and $y_i^* \in \mathcal{Y}$ represent the $i$-th question and the corresponding ground-truth answer, respectively~\cite{papadopoulos2002inductive,angelopoulos2021gentle,angelopoulos2024theoretical}. 
Generally, the COIN framework consists of three stages: \ding{182} empirical failure analysis on the calibration set, \ding{183} upper confidence bound construction, and \ding{184} threshold selection. 

\noindent\textbf{Empirical Failure Analysis on the Calibration Set.} 
In the first stage, for a given threshold $t$, we compute the empirical failure rate on the calibration set $\mathcal{D}_{cal}$. 
For each question $x_i$, we produce a candidate answer $\hat{y}_i$ and the associated uncertainty score $U\left(F\left(x_i\right) \right)$. 
Then, we define the empirical conditional failure rate (ECFR) over calibration samples whose uncertainty scores fall below $t$
\begin{equation}\label{eq: observed risk from calibration set}
    \hat{r}_{cal}\left( t \right) = \hat{w}_{cal}\left(t\right) / \hat{m}_{cal}\left(t\right),
\end{equation}
where $\hat{m}_{cal}\left(t\right)$ is is the number of selected calibration data points under the threshold $t$ 
\begin{equation}
    \hat{m}_{cal}\left(t\right) = \textstyle\sum_{\left(x_i, y_i^*\right)\in \mathcal{D}_{cal}} \mathbf{1} \left\{ U\left( F\left(x_i \right) \right) \leq t \right\},
\end{equation}
and $\hat{w}_{cal}\left(t\right)$ is the number of failures observed from the selected subset of $\mathcal{D}_{cal}$
\begin{equation}
\hat{w}_{cal}\left(t\right)=\textstyle\sum_{\left(x_i, y_i^*\right)\in \mathcal{D}_{cal}} \mathbf{1} \left\{ U\left( F\left(x_i \right) \right) \leq t \mathrm{\;and\;} A\left( \hat{y}_i, y_i^* \right) = 0 \right\}.
\end{equation}
This observed ECFR $\hat{r}_{cal}\left( t \right)$ serves as the foundation for constructing the upper confidence bound of the TCFR $R\left(t\right)$ in the second stage. 

\noindent\textbf{Upper Confidence Bound Construction.} 
To obtain a statistically rigorous guarantee of the TCFR $R\left(t\right)$, we aim to construct an upper confidence bound, denoted as $\hat{R}_{1-\delta}^{\text{upper}}$, based on the ECFR $\hat{r}_{cal}\left(t\right)$ observed from the calibration set, such that 
\begin{equation}\label{eq: probability inequality of upper bound}
\mathrm{Pr}\left( R\left(t\right) \leq \hat{R}_{1-\delta}^{\text{upper}}\left(t\right) \right) \geq 1 - \delta. 
\end{equation}
That is, with probability at least $1-\delta$ (e.g., 0.95 or 0.99), the TCFR $R\left(t\right)$ is strictly constrained by $\hat{R}_{1-\delta}^{\text{upper}}\left(t\right)$. 
To do so, we next state an auxiliary lemma~\cite{clopper1934use}.

\begin{lemma}[Clopper–Pearson Exact Confidence Interval for a Binomial Proportion]\label{lm: Clopper–Pearson}
Let \( W \sim \text{Binomial}(m, R) \), and we observe \( w \in \{0, 1, \dots, m\} \) failures. 
For any confidence level \( 1 - \delta \in (0,1) \), the Clopper–Pearson exact two-sided confidence interval\footnote{The Clopper–Pearson interval is an early and very common method for calculating binomial confidence intervals. 
It is often called an ``exact'' method, as it attains the nominal coverage level in an exact sense, meaning that the coverage level is never less than the nominal $1-\alpha$~\cite{clopper1934use,newcombe1998two}.} for \( R \) is given by:
\[
\left[ R^{\text{lower}}_{\delta/2},\ R^{\text{upper}}_{1-\delta/2} \right],
\quad \text{such that} \quad 
\mathrm{Pr} \left( R \in \left[ R^{\text{lower}}_{\delta/2}, R^{\text{upper}}_{1-\delta/2} \right] \right) \ge 1 - \delta.
\]

Each endpoint is defined via the inverse cumulative distribution function (quantile function) of the Beta distribution:
\[
R^{\text{lower}}_{\delta/2} = \text{BetaInv} \left( \frac{\delta}{2};\ w,\ m - w + 1 \right),
\quad 
R^{\text{upper}}_{1 - \delta/2} = \text{BetaInv} \left( 1 - \frac{\delta}{2};\ w + 1,\ m - w \right),
\]
where \( \text{BetaInv}(p; a, b) \) denotes the \( p \)-th quantile from a beta distribution \( \text{Beta}(a, b) \) with shape parameters $a$ and $b$~\cite{wadsworth1961introduction,thulin2013cost}.
\end{lemma}

Considering our goal of bounding the worst-case risk with high confidence, rather than estimating both ends of an interval, we adopt the one-sided variant of the Clopper–Pearson bound, which corresponds to the upper endpoint of the two-sided interval in Lemma~\ref{lm: Clopper–Pearson}. Specifically, given the observed number of failures \( w \) among \( m \) i.i.d. Bernoulli trials with the true failure rate $R$, the one-sided upper bound of $R$ with confidence level \( 1 - \delta \) is defined as:
\begin{equation}
    R^{\text{upper}}_{1 - \delta} = \text{BetaInv}(1 - \delta;\ w + 1,\ m - w),
\end{equation}
which satisfies $
\mathbb{P} \left( R \le R^{\text{upper}}_{1 - \delta} \right) \ge 1 - \delta.$

In our setting, for an uncertainty threshold \( t \), \( \hat{m}_{\text{cal}}(t) \) denotes the number of selected calibration data points and \( \hat{w}_{\text{cal}}(t) \) is the number of failures among them. 
Then, the one-sided Clopper–Pearson exact upper confidence bound on the TCFR \( R(t) \) is given by:
\begin{equation}\label{eq: beta upper bound}
    \hat{R}^{\text{upper}}_{1-\delta}(t) = \text{BetaInv} \left( 1 - \delta;\ \hat{w}_{\text{cal}}(t) + 1,\ \hat{m}_{\text{cal}}(t) - \hat{w}_{\text{cal}}(t) \right).
\end{equation}

While the bound \( \hat{R}^{\text{upper}}_{1 - \delta}(t) \) in Eq.~\ref{eq: beta upper bound} provides a closed-form expression via the Beta inverse cumulative distribution function (CDF)\footnote{The bound \( \hat{R}^{\text{upper}}_{1-\delta}(t) \) can be computed using standard Beta inverse CDF functions implemented in statistical libraries such as \texttt{scipy.stats.beta.ppf} or \texttt{R::qbeta}.}, it is equally instructive to reinterpret this bound through the lens of the underlying Bernoulli model. In particular, we analyze the distributional behavior of the random empirical failure rate under the true but unknown conditional failure rate \( R(t) \). This alternative view enables us to formulate \( \hat{R}^{\text{upper}}_{1 - \delta}(t) \) directly from the cumulative distribution of the empirical estimator.

Under the Bernoulli structure, the number of observed failures among \( \hat{m}_{\text{cal}}(t) \) selected data points follows a Binomial distribution with success probability \( R(t) \). 
We then denote this random variable as \( W \sim \text{Binomial}(\hat{m}_{\text{cal}}(t), R(t)) \), and define the random empirical failure rate as $\hat{R}\left(t\right) = \mathrm{Binomial }\left( \hat{m}_{cal}\left(t\right), R\left(t\right) \right)/{\hat{m}_{cal}\left(t\right)}$. That is, \( \hat{R}(t) \) is a scaled Binomial random variable representing the empirical frequency of failures among selected samples under threshold \( t \), and $\hat{r}_{cal}\left( t \right)$ defined in Eq~\ref{eq: observed risk from calibration set} is its realization observed from the calibration set. 
To formally express the statistical confidence constraint, we introduce the CDF \( D(r \mid R(t)) \) of \( \hat{R}(t) \) under the true failure rate \( R(t) \):
\begin{equation}\label{eq: cdf}
    \begin{split}
    D\left(r \mid R\left( t \right) \right) &= \mathrm{Pr}\left( \hat{R}\left(t\right) \leq r \mid R\left( t \right) \right)\\
\end{split}.
\end{equation}
This CDF characterizes the likelihood that the empirical failure rate is no greater than a specified risk threshold \( r \), under the true distribution governed by \( R(t) \).

With this setup, we can reinterpret the Clopper–Pearson upper confidence bound as the largest value of \( R(t) \in [0, 1] \) such that the observed empirical failure rate \( \hat{r}_{\text{cal}}(t) \) remains a statistically plausible realization with probability at least \( \delta \). 
This leads to the equivalent formulation:
\begin{equation}\label{eq: upper bound via CDF}
    \hat{R}^{\text{upper}}_{1 - \delta}(t) = \sup \left\{ R(t) \in [0,1] : D(\hat{r}_{\text{cal}}(t) \mid R(t)) \ge \delta \right\},
\end{equation}
which unifies both the exact and approximate constructions. 
\textbf{Intuitively}, since the ECFR $\hat{r}_{cal}\left(t\right)$ has already been observed on the calibration set, it cannot correspond to a rare (small probability) event under the TCFR $R\left( t \right)$, which concludes the high-probability (i.e., $1-\delta$) upper bound $\hat{R}_{1-\delta}^{\text{upper}}\left(t\right)$. 
A statistical proof that $\hat{R}^{\text{upper}}_{1 - \delta}(t)$ in Eq.~\ref{eq: upper bound via CDF} satisfies the guarantee in Eq.~\ref{eq: probability inequality of upper bound} is provided in Appendix~\ref{appendix: proofs}. 

\noindent\textbf{Threshold Selection.} 
Since the TCFR $R\left(t\right)$ is typically non-decreasing with respect to the uncertainty threshold $t$, we exploit this monotonicity to identify the largest threshold that satisfies the risk constraint. 
Formally, we define the calibrated threshold \( \hat{t} \) as:
\begin{equation}
    \hat{t} = \sup \left\{ t: \hat{R}^{\text{upper}}_{1-\delta}(t') \le \alpha \text{ for all } t' \le t \right\}.
\end{equation}
We select the largest such threshold to retain the most candidate answers while ensuring that the associated upper confidence bound $\hat{R}_{1-\delta}^{\text{upper}}\left(t\right)$ does not exceed the target risk level $\alpha$.
Moreover, this construction circumvents the need for conservative multiple hypothesis testing corrections, such as the Bonferroni adjustment, which would otherwise be required if threshold selection is performed post hoc without valid uniform coverage guarantees~\cite{angelopoulos2021learn,quach2024conformal,gui2024conformal}. 
Finally, the calibrated acceptance policy satisfies
\begin{equation}
    \mathrm{Pr}\left( R(\hat{t}) \le \alpha \right) \ge 1 - \delta,
\end{equation}
thus ensuring that, with confidence at least \( 1 - \delta \), the TCFR of the selected model generations is guaranteed to stay below the user-specified risk level \( \alpha \).

\subsection{Implementation Practicality}
\noindent\textbf{Clopper–Pearson Exact Upper Confidence Bound.} 
In practice, for a candidate threshold $t$, computing the upper bound \( \hat{R}^{\text{upper}}_{1-\delta}(t) \) reduces to solving a one-dimensional root-finding problem: identifying the largest value of \( R(t) \in [0, 1] \) such that the probability of observing at most \( \hat{w}_{\text{cal}}(t) \) failures in \( \hat{m}_{\text{cal}}(t) \) Bernoulli trials is at least \( \delta \), i.e., $\mathrm{Pr}(\text{Binomial}(\hat{m}_{\text{cal}}(t), R(t)) \leq \hat{w}_{\text{cal}}(t)) \geq \delta$.  
Since the binomial CDF is monotonic in \( R(t) \), this optimization can be efficiently solved via binary search over the interval \( R(t) \in [\hat{r}_{\text{cal}}(t), 1] \). 
The procedure does not require variance estimation or asymptotic approximation, making it robust and stable even in deployment-critical environments.

\begin{wrapfigure}{R}{0.51\textwidth}
    \centering    
\includegraphics[width=0.51\textwidth]{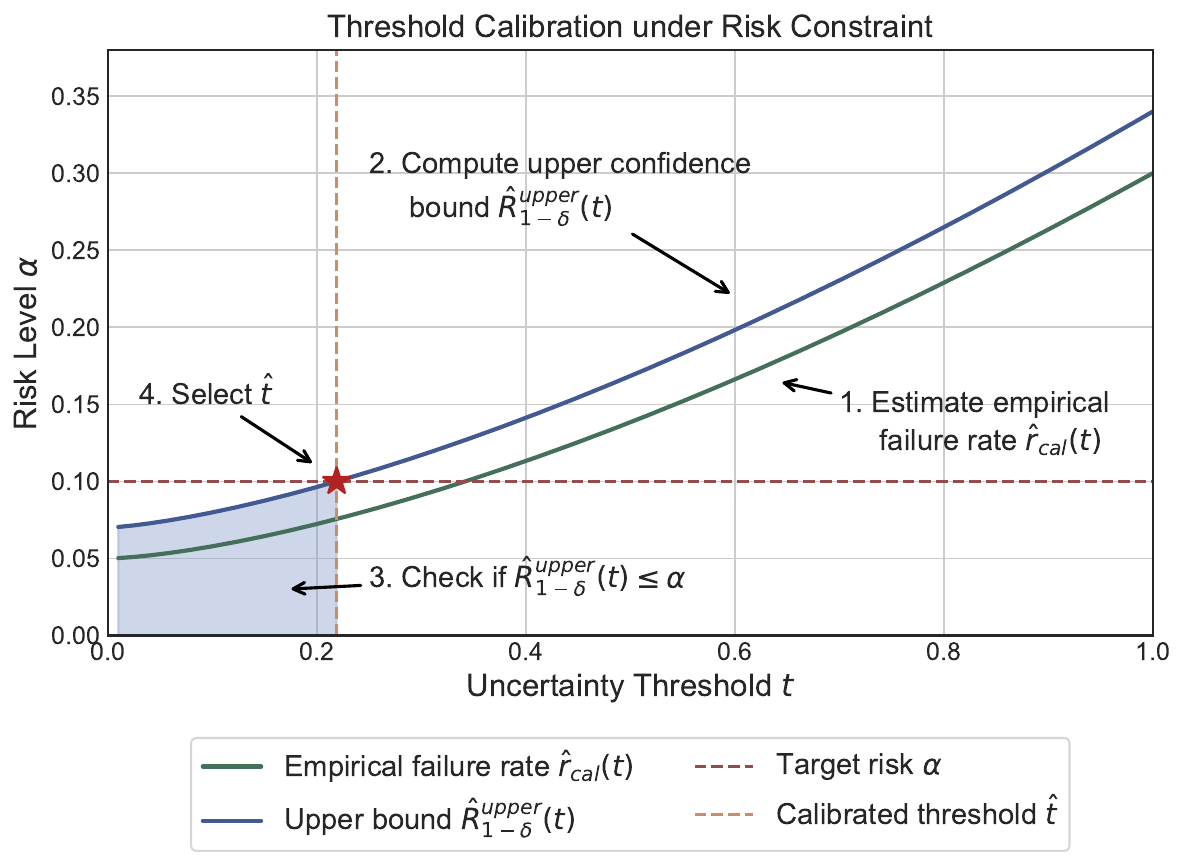}
    \caption{An illustration of the threshold calibration. Once computing the ECFR on selected calibration samples given a candidate threshold $t$, we construct the upper confidence bound of the TCFR. 
    We select the largest $t$, i.e., $\hat{t}$, while ensuring that the upper bound falls below the risk level $\alpha$.}\label{fig: implementation}
\end{wrapfigure}

\noindent\textbf{Threshold Selection.} 
As illustrated in Figure~\ref{fig: implementation}, 
given the computed values of \( \hat{R}^{\text{upper}}_{1 - \delta}(t) \), we select the largest threshold \( t \) (i.e., $\hat{t}$) such that the risk constraint \( \hat{R}^{\text{upper}}_{1 - \delta}(t) \le \alpha \) is satisfied.
This selection procedure exploits the empirical near-monotonicity of the TCFR \( R(t) \) for \( t \), thereby enabling us to maximize the retained samples while certifying that the final selection policy respects the desired risk level.

\noindent\textbf{Concentration-Based Alternatives.} 
While the Clopper–Pearson interval provides exact and non-asymptotic guarantees, it tends to incur high computational costs when evaluated over a large number of uncertainty thresholds for each given risk level~\cite{chung2006concentration,thulin2013cost}. 
To improve scalability, we explore alternative approaches that yield approximate yet statistically valid intervals, particularly suitable for large-scale and computationally constrained scenarios. 
As a representative example, we adopt a concentration-based upper bound derived from Hoeffding’s inequality~\cite{hoeffding1994probability,jo2021machine,bates2021distribution,angelopoulos2021learn}, which yields a distribution-free upper confidence bound on the TCFR \( R(t) \), relying solely on the empirical failure rate \( \hat{r}_{\text{cal}}(t) \) and the number of selected samples \( \hat{m}_{\text{cal}}(t) \). 
We next present an auxiliary lemma. 

\begin{lemma}[Hoeffding’s Inequality for Bernoulli Random Variables]\label{lm: Hoeffding-style Upper Confidence Bound}
    Let \( W_1, \ldots, W_m \in \{0,1\} \) be independent Bernoulli random variables with common mean \( \mu = \mathbb{E}\left[W_i\right] \in [0,1] \). 
    Then for any $t\geq0$, the following inequalities hold:
    \begin{equation}\label{eq: brv Hoeffding’s inequality}
        \mathrm{Pr} \left( \left| \frac{1}{m} \sum_{i=1}^{m} W_i - \mu \right| \geq t \right) \leq 2 \exp \left(-2mt^2\right).
    \end{equation}
    Equivalently, with probability at least $1-\delta$ ($\delta=\exp \left(-2mt^2\right)$), the population mean satisfies 
    \begin{equation}\label{eq: lemma 2 one-sided bound}
        \mu \leq \frac{1}{m} \sum_{i=1}^m W_i + \sqrt{\frac{1}{2m} \log \frac{1}{\delta}}.
    \end{equation}
\end{lemma}

Let $m = \hat{m}_{\text{cal}}(t)$ and $\hat{r}_{\text{cal}}(t) = \frac{1}{\hat{m}_{\text{cal}}(t)} \sum_{i=1}^{\hat{m}_{\text{cal}}(t)} W_i$. 
By applying Lemma 2, we obtain a distribution-free, closed-form upper confidence bound on the TCFR \( R(t) \) as:
\begin{equation}\label{eq: upper-h}
    \hat{R}^{\text{upper-H}}_{1-\delta}(t) = \hat{r}_{\text{cal}}(t) + \sqrt{\frac{1}{2 \hat{m}_{\text{cal}}(t)} \log \frac{1}{\delta}}.
\end{equation}

We then define the Hoeffding-style risk-bounded threshold as:
\begin{equation}
    \hat{t}^{\text{H}} = \sup \left\{ t  : \hat{R}^{\text{upper-H}}_{1-\delta}(t') \le \alpha \text{ for all } t' \le t \right\}.
\end{equation}
This yields a lightweight, closed-form alternative to Clopper–Pearson-based calibration while maintaining a valid high-probability control over the TCFR. 
A complete proof of Lemma~\ref{lm: Hoeffding-style Upper Confidence Bound} is provided in Appendix~\ref{appendix: proofs}. 
Together, these constructions provide a complete and computationally feasible procedure for risk-controlled threshold selection. 
In the next section, the COIN frameworks using the two upper-bound construction methods are denoted as COIN-CP and COIN-HFD, respectively.

\begin{figure*}[!t]
    \centering
    \begin{subfigure}{0.495\linewidth}
        \centering
        \includegraphics[width=\linewidth]{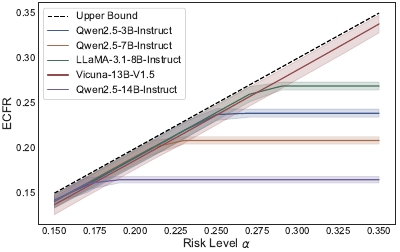}
        \caption{CommonsenseQA.}
	\label{fig: risk control mcqa Clopper–Pearson}
    \end{subfigure}
    \hfill
    \centering
    \begin{subfigure}{0.495\linewidth}
	\centering
	\includegraphics[width=\linewidth]{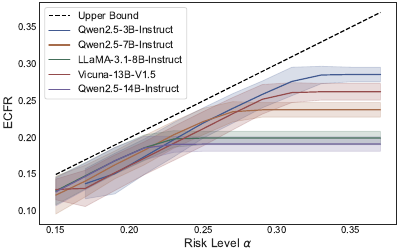}
	\caption{TriviaQA.}
	\label{fig: risk control open-domain qa Clopper–Pearson}
    \end{subfigure}
    \caption{Results of the ECFR on the test set (i.e., FDR), implementing COIN-CP, on both CommonsenseQA and TriviaQA datasets utilizing five LLMs of sizes ranging from 3B to 14B.}\label{fig: ECFR llm Clopper–Pearson}
\end{figure*}

\begin{figure*}[!t]
    \centering
    \begin{subfigure}{0.495\linewidth}
        \centering
        \includegraphics[width=\linewidth]{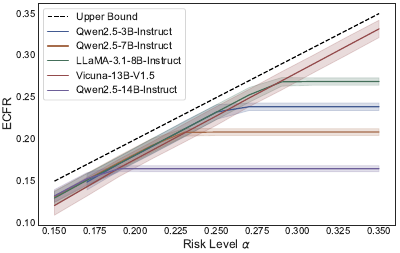}
        \caption{CommonsenseQA.}
	\label{fig: risk control mcqa Hoeffding}
    \end{subfigure}
    \hfill
    \centering
    \begin{subfigure}{0.495\linewidth}
	\centering
	\includegraphics[width=\linewidth]{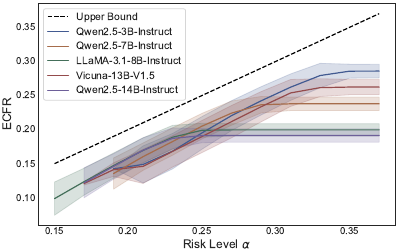}
	\caption{TriviaQA.}
	\label{fig: risk control open-domain qa Hoeffding}
    \end{subfigure}
    \caption{Results of the ECFR on the test set (i.e., FDR), implementing COIN-HFD, on both CommonsenseQA and TriviaQA datasets utilizing five LLMs of sizes ranging from 3B to 14B.}\label{fig: ECFR llm Hoeffding}
\end{figure*}

\section{Experiments}
\subsection{Experimental Settings}\label{sec: Experimental Settings}
\noindent\textbf{Datasets and Models.} 
We evaluate COIN on the closed-ended CommonsenseQA dataset~\cite{talmor-etal-2019-commonsenseqa} and the open-domain TriviaQA dataset~\cite{joshi-etal-2017-triviaqa} across five LLMs. 
Following \cite{zhang2024vl}, we consider the open-domain multimodal QA dataset: MMVet~\cite{yu2024mmvet}, employing five LVLMs. 
More details can be found in Appendix~\ref{appendix: Details of Datasets} and Appendix~\ref{appendix: Details of Models}. 


\noindent\textbf{Uncertainty Measures.} 
COIN accommodates various UQ methods at both white-box and black-box settings. 
By default, we adopt predictive entropy (\textit{PE})~\cite{kadavath2022language} for closed-ended QA tasks, and semantic entropy (\textit{SE})~\cite{kuhn2023semantic,farquhar2024detecting,zhang2024vl} for open-domain QA tasks.
Additionally, we incorporate three UQ techniques: \textit{Ecc}, \textit{Deg}, and \textit{Eigv}, introduced in \cite{lingenerating}. 
More implementation details can be found in Appendix~\ref{appendix: Details of Uncertainty Measures}. 

\noindent\textbf{Admission Function.} 
Following prior work~\cite{wang2025word,wang2025sconu}, we employ sentence similarity as the correctness metric in open-domain scenarios by default, where the model output is deemed as correct if it has above 0.7 semantic similarities with the ground-truth, measured by SentenceTransformers~\cite{reimers-gurevych-2019-sentence} with DistillRoBERTa~\cite{sanh2019distilbert} as the backbone. 
We also consider bi-entailment and LLM judgment. 
More details can be found in Appendix~\ref{appendix: Details of Correctness Metrics}. 

\noindent\textbf{Evaluation Metrics.} 
We examine the ECFR on the test set (i.e., FDR) to evaluate the statistical validity of COIN. 
Following CA~\cite{gui2024conformal}, we also define the power metric as the proportion of admissible answers in the test set that are selected. 
Disregarding the model's overall performance, our goal is to select as many correct answers as possible while maintaining risk control. 
 
\noindent\textbf{Hyperparameters.} 
Following~\cite{duan2024shifting,wang2025word}, we leverage beam search (\texttt{num\_beams=5}) for the most likely generation, which is then employed as the model output. 
For closed-ended QA tasks, given that we develop prompts to guide the language model in responding with the most probable option (e.g., A, B, or C), we set the maximum generation length to 1. 
For open-domain QA tasks, we set the maximum generation length to 32. 
We set the significance level $\delta$ to 0.05. 
In addition, we fix the split ratio between the calibration and test data to 0.5. 

\begin{figure*}[!t]
    \centering
    \begin{subfigure}{0.495\linewidth}
        \centering
        \includegraphics[width=\linewidth]{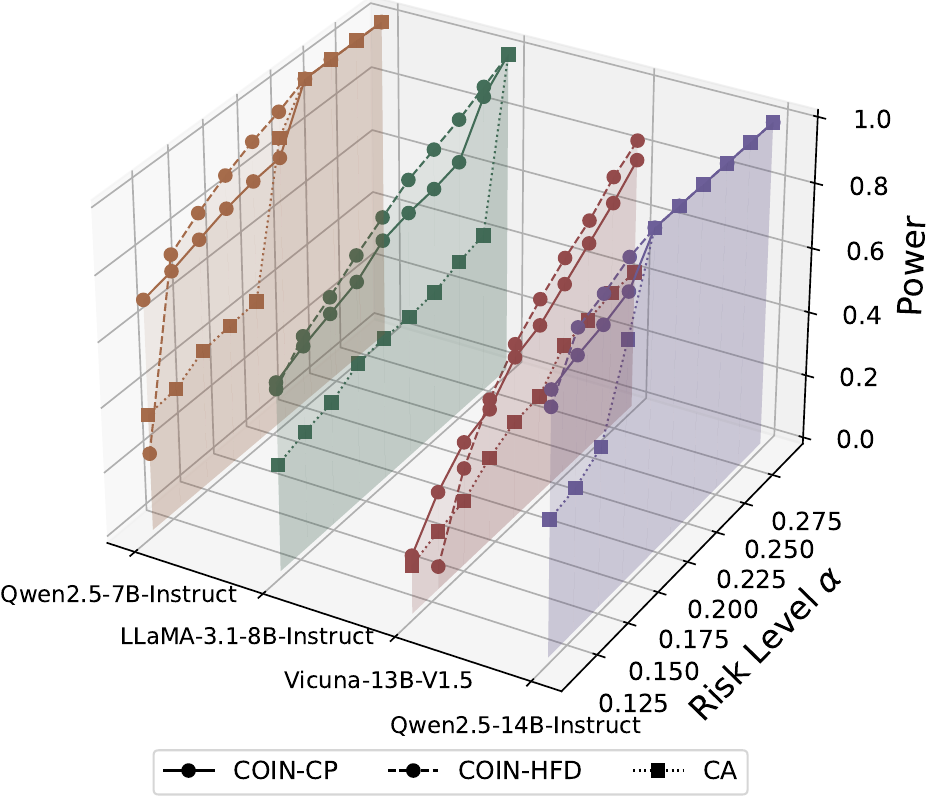}
        \caption{CommonsenseQA.}
	\label{fig: power mcqa}
    \end{subfigure}
    \hfill
    \centering
    \begin{subfigure}{0.495\linewidth}
	\centering
	\includegraphics[width=\linewidth]{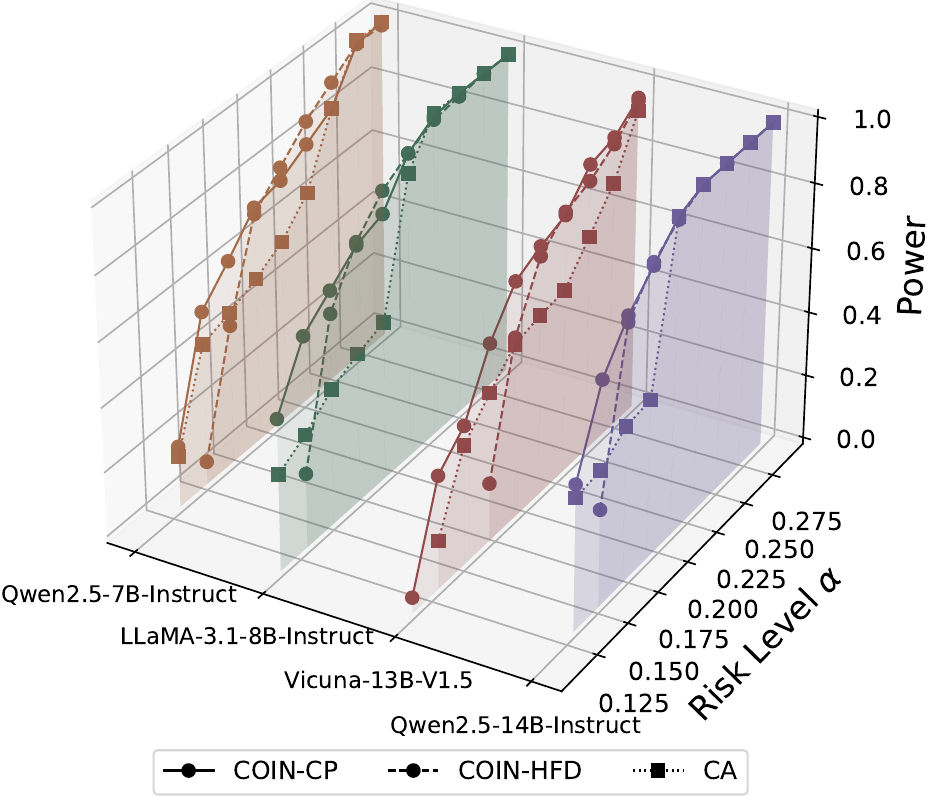}
	\caption{TriviaQA.}
	\label{fig: power open-domain qa}
    \end{subfigure}
    \caption{Comparison of the power metric on both textual QA datasets across 4 LLMs. COIN-CP consistently outperforms CA at various user-specified risk levels.}\label{fig: power}
\end{figure*}

\subsection{Empirical Evaluations}\label{sec: Empirical Evaluations}
\paragraph{Statistical Validity of COIN.} 
We evaluate COIN-CP and COIN-HFD on the CommonsenseQA and TriviaQA benchmarks.
The ECFR results across 5 LLMs are presented in Figure~\ref{fig: ECFR llm Clopper–Pearson} and Figure~\ref{fig: ECFR llm Hoeffding}.
Each solid line represents the mean ECFR over 100 trials and the shaded areas indicate $+/-$ the standard deviation. 
Both COIN-CP and COIN-HFD rigorously constrain the test-time ECFR below a range of user-specified risk levels. 
Considering that the Clopper–Pearson method yields an exact (non-asymptotic) confidence interval for binomial proportions, whereas the Hoeffding-based upper bound, derived from concentration inequalities, is typically looser~\cite{bates2021distribution,angelopoulos2021learn}, the ECFR achieved by COIN-CP tends to be closely concentrated just below the upper bound, while the ECFR of COIN-HFD remains further below the target risk level. 

Additional results of ECFR on the multimodal MMVet dataset, as well as those obtained leveraging other UQ methods under both the white-box and black-box settings, alternative correctness evaluation criteria, and varying calibration-to-test set ratios, are reported in Appendix~\ref{appendix: Additional Experimental Results}. 

\paragraph{Comparison of Power.} 
Without compromising risk control, we aim to include as many admissible answers as possible from new samples.
As demonstrated in Figure~\ref{fig: power}, COIN-CP consistently outperforms CA across a range of desired risk levels.
For instance, on the CommonsenseQA dataset with the Qwen-2.5-14B-Instruct model, COIN-CP achieves a power of 0.85 at a relatively low risk level of 0.15, exceeding CA by 0.36, and reaches full power (1) at a risk level of only 0.19. 
Moreover, COIN-CP can surpass CA by up to 0.4 in power at a risk level of 0.19 on the TriviaQA dataset. 
Furthermore, when adopting the Hoeffding-style upper bound, COIN-HFD generally yields better performance than COIN-CP on CommonsenseQA. 
For example, employing the Vicuna-13B-V1.5 model, COIN-HFD achieves a power of 0.92 at a risk level of 0.25, 0.13 higher than COIN-CP and 0.44 higher than CA. 
These results demonstrate that optimizing the upper bound construction in the second stage can significantly enhance the power of COIN while constraining the risk.

Note that all reported power results are averaged over 100 trials. For each trial, the power is counted only if the ECFR on the test set falls below the specified risk level; otherwise, it is recorded as zero. 
We also discuss the improvements in power performance when using alternative UQ methods in the first stage in Appendix~\ref{appendix: Additional Experimental Results}. 
This underscores the modularity and extensibility of the COIN framework, where each stage can be independently tuned without sacrificing theoretical guarantees. 
Such flexibility enables COIN to adapt to diverse application needs and computational constraints, facilitating task-specific trade-offs between statistical rigor and practical performance.

\subsection{Sensitivity Analysis}
\begin{figure*}[!t]
    \centering
    \begin{subfigure}{0.495\linewidth}
        \centering
        \includegraphics[width=\linewidth]{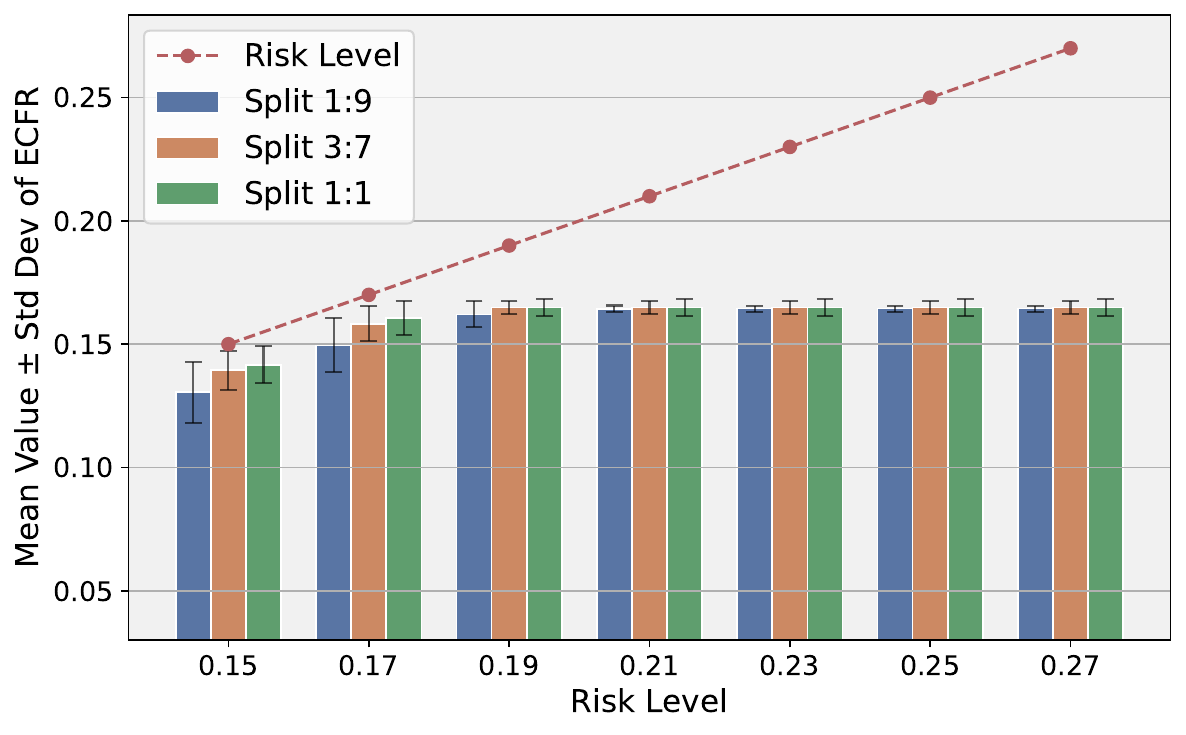}
        \caption{CommonsenseQA.}
	\label{fig: commonsenseqa split ratios}
    \end{subfigure}
    \hfill
    \centering
    \begin{subfigure}{0.495\linewidth}
	\centering
	\includegraphics[width=\linewidth]{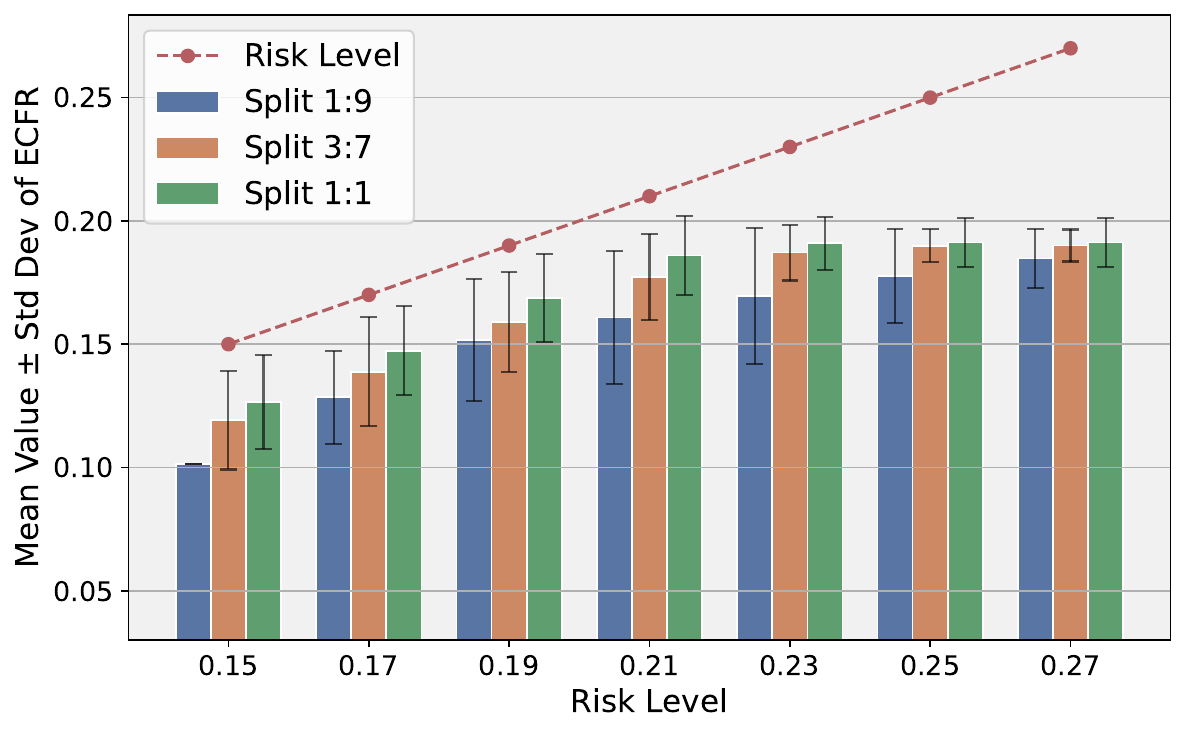}
	\caption{TriviaQA.}
	\label{fig: triviaqa split ratios}
    \end{subfigure}
    \caption{Results of the ECFR on the test set on both textual QA datasets across various calibration-to-test set ratios, employing the Qwen2.5-14B-Instruct model.}\label{fig: qa split ratios}
\end{figure*}

As previously described, COIN calibrates the uncertainty threshold using a held-out calibration set. 
We further examine whether COIN can maintain risk control on new samples with limited calibration data. 
As shown in Figure~\ref{fig: qa split ratios}, even with a 1:9 calibration-to-test data ratio, COIN reliably controls the FDR across a range of user-specified risk levels, which highlights the robustness and predictive efficiency of COIN. 
We also discuss in Appendix~\ref{appendix: Additional Experimental Results} that COIN can maintain risk control under black-box conditions when computing uncertainty scores with a smaller sampling size.

\section{Conclusion}
In this work, we present COIN, a modular three-stage framework that calibrates a statistically valid threshold to filter new samples for each user-specified risk level. 
We validate the robustness of COIN in controlling the FDR across two general QA datasets and one multimodal QA dataset. 
Compared to CA, COIN eliminates the need to train an auxiliary model. 
By pre-calibrating the maximum uncertainty threshold, COIN enables faster test-time selection and achieves higher power in retaining admissible answers. 
Moreover, COIN exhibits strong extensibility, as each stage can be independently adapted and optimized based on the task type and application scenario. 
(1) In the first stage, COIN selects the optimal white-box or black-box UQ method based on the availability of access to the model's internal information to obtain more accurate error estimations on the calibration set. 
(2) In the second stage, we can either employ the exact Clopper–Pearson upper confidence bound or adopt concentration inequality-based methods such as Hoeffding’s inequality, which provide a lower computational cost. 
(3) In the third stage, we can select the largest threshold to accept more correct samples while preserving the risk constraint. 
Furthermore, we demonstrate that COIN maintains its statistical rigor under two challenging conditions: (1) when only limited calibration data is available; (2) under black-box settings where a small sampling size is used to estimate uncertainty, which highlights its predictive efficiency. 
We hope COIN can be further applied to other downstream tasks to improve the reliability of uncertainty-aware decisions made by foundation models.



\bibliography{iclr2025_conference}

\begin{thebibliography}{71}
\providecommand{\natexlab}[1]{#1}
\providecommand{\url}[1]{\texttt{#1}}
\expandafter\ifx\csname urlstyle\endcsname\relax
  \providecommand{\doi}[1]{doi: #1}\else
  \providecommand{\doi}{doi: \begingroup \urlstyle{rm}\Url}\fi

\bibitem[Abbasli et~al.(2025)Abbasli, Toyoda, Wang, Witt, Ali, Miao, Li, and Wei]{abbasli2025comparing}
Toghrul Abbasli, Kentaroh Toyoda, Yuan Wang, Leon Witt, Muhammad~Asif Ali, Yukai Miao, Dan Li, and Qingsong Wei.
\newblock Comparing uncertainty measurement and mitigation methods for large language models: A systematic review.
\newblock \emph{arXiv preprint arXiv:2504.18346}, 2025.

\bibitem[AI@Meta(2024)]{llama3modelcard}
AI@Meta.
\newblock Llama 3 model card.
\newblock 2024.

\bibitem[Angelopoulos \& Bates(2021)Angelopoulos and Bates]{angelopoulos2021gentle}
Anastasios~N Angelopoulos and Stephen Bates.
\newblock A gentle introduction to conformal prediction and distribution-free uncertainty quantification.
\newblock \emph{arXiv preprint arXiv:2107.07511}, 2021.

\bibitem[Angelopoulos et~al.(2021)Angelopoulos, Bates, Cand{\`e}s, Jordan, and Lei]{angelopoulos2021learn}
Anastasios~N Angelopoulos, Stephen Bates, Emmanuel~J Cand{\`e}s, Michael~I Jordan, and Lihua Lei.
\newblock Learn then test: Calibrating predictive algorithms to achieve risk control.
\newblock \emph{arXiv preprint arXiv:2110.01052}, 2021.

\bibitem[Angelopoulos et~al.(2024)Angelopoulos, Barber, and Bates]{angelopoulos2024theoretical}
Anastasios~N Angelopoulos, Rina~Foygel Barber, and Stephen Bates.
\newblock Theoretical foundations of conformal prediction.
\newblock \emph{arXiv preprint arXiv:2411.11824}, 2024.

\bibitem[Bates et~al.(2021)Bates, Angelopoulos, Lei, Malik, and Jordan]{bates2021distribution}
Stephen Bates, Anastasios Angelopoulos, Lihua Lei, Jitendra Malik, and Michael Jordan.
\newblock Distribution-free, risk-controlling prediction sets.
\newblock \emph{Journal of the ACM (JACM)}, 2021.

\bibitem[Batu et~al.(2001)Batu, Fischer, Fortnow, Kumar, Rubinfeld, and White]{batu2001testing}
Tugkan Batu, Eldar Fischer, Lance Fortnow, Ravi Kumar, Ronitt Rubinfeld, and Patrick White.
\newblock Testing random variables for independence and identity.
\newblock In \emph{Proceedings 42nd IEEE Symposium on Foundations of Computer Science}, 2001.

\bibitem[Benjamini \& Hochberg(1995)Benjamini and Hochberg]{benjamini1995controlling}
Yoav Benjamini and Yosef Hochberg.
\newblock Controlling the false discovery rate: a practical and powerful approach to multiple testing.
\newblock \emph{Journal of the Royal statistical society: series B (Methodological)}, 1995.

\bibitem[Bentkus(2004)]{bentkus2004hoeffding}
Vidmantas Bentkus.
\newblock On hoeffding's inequalities.
\newblock \emph{Annals of Probability}, pp.\  1650--1673, 2004.

\bibitem[Campos et~al.(2024)Campos, Farinhas, Zerva, Figueiredo, and Martins]{campos-etal-2024-conformal}
Margarida Campos, Ant{\'o}nio Farinhas, Chrysoula Zerva, M{\'a}rio A.~T. Figueiredo, and Andr{\'e} F.~T. Martins.
\newblock Conformal prediction for natural language processing: A survey.
\newblock \emph{Transactions of the Association for Computational Linguistics}, 2024.

\bibitem[Chen et~al.(2023)Chen, Yoon, Ebrahimi, Arik, Pfister, and Jha]{chen-etal-2023-adaptation}
Jiefeng Chen, Jinsung Yoon, Sayna Ebrahimi, Sercan Arik, Tomas Pfister, and Somesh Jha.
\newblock Adaptation with self-evaluation to improve selective prediction in {LLM}s.
\newblock In \emph{Findings of the Association for Computational Linguistics: EMNLP 2023}, 2023.

\bibitem[Chen et~al.(2024)Chen, Wu, Wang, Su, Chen, Xing, Zhong, Zhang, Zhu, Lu, et~al.]{chen2024internvl}
Zhe Chen, Jiannan Wu, Wenhai Wang, Weijie Su, Guo Chen, Sen Xing, Muyan Zhong, Qinglong Zhang, Xizhou Zhu, Lewei Lu, et~al.
\newblock Internvl: Scaling up vision foundation models and aligning for generic visual-linguistic tasks.
\newblock In \emph{Proceedings of the IEEE/CVF Conference on Computer Vision and Pattern Recognition}, 2024.

\bibitem[Cherian et~al.(2024)Cherian, Gibbs, and Candes]{cherian2024large}
John Cherian, Isaac Gibbs, and Emmanuel Candes.
\newblock Large language model validity via enhanced conformal prediction methods.
\newblock \emph{Advances in Neural Information Processing Systems}, 2024.

\bibitem[Chung \& Lu(2006)Chung and Lu]{chung2006concentration}
Fan Chung and Linyuan Lu.
\newblock Concentration inequalities and martingale inequalities: a survey.
\newblock \emph{Internet mathematics}, 2006.

\bibitem[Clopper \& Pearson(1934)Clopper and Pearson]{clopper1934use}
Charles~J Clopper and Egon~S Pearson.
\newblock The use of confidence or fiducial limits illustrated in the case of the binomial.
\newblock \emph{Biometrika}, 1934.

\bibitem[Cresswell et~al.(2025)Cresswell, Kumar, Sui, and Belbahri]{Jesse2025Disparate}
Jesse~C. Cresswell, Bhargava Kumar, Yi~Sui, and Mouloud Belbahri.
\newblock Conformal prediction sets can cause disparate impact.
\newblock In \emph{The Thirteenth International Conference on Learning Representations}, 2025.

\bibitem[Duan et~al.(2024)Duan, Cheng, Wang, Zavalny, Wang, Xu, Kailkhura, and Xu]{duan2024shifting}
Jinhao Duan, Hao Cheng, Shiqi Wang, Alex Zavalny, Chenan Wang, Renjing Xu, Bhavya Kailkhura, and Kaidi Xu.
\newblock Shifting attention to relevance: Towards the predictive uncertainty quantification of free-form large language models.
\newblock In \emph{Proceedings of the 62nd Annual Meeting of the Association for Computational Linguistics}, 2024.

\bibitem[Duan et~al.(2025)Duan, Kong, Cheng, Diffenderfer, Kailkhura, Sun, Zhu, Shi, and Xu]{duan2025truthprint}
Jinhao Duan, Fei Kong, Hao Cheng, James Diffenderfer, Bhavya Kailkhura, Lichao Sun, Xiaofeng Zhu, Xiaoshuang Shi, and Kaidi Xu.
\newblock Truthprint: Mitigating lvlm object hallucination via latent truthful-guided pre-intervention.
\newblock \emph{arXiv preprint arXiv:2503.10602}, 2025.

\bibitem[Farquhar et~al.(2024)Farquhar, Kossen, Kuhn, and Gal]{farquhar2024detecting}
Sebastian Farquhar, Jannik Kossen, Lorenz Kuhn, and Yarin Gal.
\newblock Detecting hallucinations in large language models using semantic entropy.
\newblock \emph{Nature}, 2024.

\bibitem[Gui et~al.(2024)Gui, Jin, and Ren]{gui2024conformal}
Yu~Gui, Ying Jin, and Zhimei Ren.
\newblock Conformal alignment: Knowing when to trust foundation models with guarantees.
\newblock In \emph{The Thirty-eighth Annual Conference on Neural Information Processing Systems}, 2024.

\bibitem[Guo et~al.(2025)Guo, Yang, Zhang, Song, Zhang, Xu, Zhu, Ma, Wang, Bi, et~al.]{guo2025deepseek}
Daya Guo, Dejian Yang, Haowei Zhang, Junxiao Song, Ruoyu Zhang, Runxin Xu, Qihao Zhu, Shirong Ma, Peiyi Wang, Xiao Bi, et~al.
\newblock Deepseek-r1: Incentivizing reasoning capability in llms via reinforcement learning.
\newblock \emph{arXiv preprint arXiv:2501.12948}, 2025.

\bibitem[He et~al.(2021)He, Liu, Gao, and Chen]{he2021deberta}
Pengcheng He, Xiaodong Liu, Jianfeng Gao, and Weizhu Chen.
\newblock Deberta: Decoding-enhanced bert with disentangled attention.
\newblock In \emph{International Conference on Learning Representations}, 2021.

\bibitem[Hoeffding(1994)]{hoeffding1994probability}
Wassily Hoeffding.
\newblock Probability inequalities for sums of bounded random variables.
\newblock \emph{The collected works of Wassily Hoeffding}, 1994.

\bibitem[Hou et~al.(2025)Hou, Zhang, Andreas, and Chang]{hou-etal-2025-probabilistic}
Bairu Hou, Yang Zhang, Jacob Andreas, and Shiyu Chang.
\newblock A probabilistic framework for {LLM} hallucination detection via belief tree propagation.
\newblock In \emph{Proceedings of the 2025 Conference of the Nations of the Americas Chapter of the Association for Computational Linguistics: Human Language Technologies (Volume 1: Long Papers)}, 2025.

\bibitem[Huang et~al.(2025)Huang, Yu, Ma, Zhong, Feng, Wang, Chen, Peng, Feng, Qin, et~al.]{huang2025survey}
Lei Huang, Weijiang Yu, Weitao Ma, Weihong Zhong, Zhangyin Feng, Haotian Wang, Qianglong Chen, Weihua Peng, Xiaocheng Feng, Bing Qin, et~al.
\newblock A survey on hallucination in large language models: Principles, taxonomy, challenges, and open questions.
\newblock \emph{ACM Transactions on Information Systems}, 2025.

\bibitem[Huang et~al.(2024)Huang, Lala, and Jha]{huang2024confine}
Linhui Huang, Sayeri Lala, and Niraj~K Jha.
\newblock Confine: Conformal prediction for interpretable neural networks.
\newblock \emph{arXiv preprint arXiv:2406.00539}, 2024.

\bibitem[Hurst et~al.(2024)Hurst, Lerer, Goucher, Perelman, Ramesh, Clark, Ostrow, Welihinda, Hayes, Radford, et~al.]{hurst2024gpt}
Aaron Hurst, Adam Lerer, Adam~P Goucher, Adam Perelman, Aditya Ramesh, Aidan Clark, AJ~Ostrow, Akila Welihinda, Alan Hayes, Alec Radford, et~al.
\newblock Gpt-4o system card.
\newblock \emph{arXiv preprint arXiv:2410.21276}, 2024.

\bibitem[Jin \& Cand{\`e}s(2023)Jin and Cand{\`e}s]{jin2023selection}
Ying Jin and Emmanuel~J Cand{\`e}s.
\newblock Selection by prediction with conformal p-values.
\newblock \emph{Journal of Machine Learning Research}, 2023.

\bibitem[Jin \& Ren(2024)Jin and Ren]{jin2024confidence}
Ying Jin and Zhimei Ren.
\newblock Confidence on the focal: Conformal prediction with selection-conditional coverage.
\newblock \emph{arXiv preprint arXiv:2403.03868}, 2024.

\bibitem[Jo(2021)]{jo2021machine}
Taeho Jo.
\newblock Machine learning foundations.
\newblock \emph{Supervised, Unsupervised, and Advanced Learning. Cham: Springer International Publishing}, 2021.

\bibitem[Joshi et~al.(2017)Joshi, Choi, Weld, and Zettlemoyer]{joshi-etal-2017-triviaqa}
Mandar Joshi, Eunsol Choi, Daniel Weld, and Luke Zettlemoyer.
\newblock {T}rivia{QA}: A large scale distantly supervised challenge dataset for reading comprehension.
\newblock In \emph{Proceedings of the 55th Annual Meeting of the Association for Computational Linguistics}, 2017.

\bibitem[Kadavath et~al.(2022)Kadavath, Conerly, Askell, Henighan, Drain, Perez, Schiefer, Hatfield-Dodds, DasSarma, Tran-Johnson, et~al.]{kadavath2022language}
Saurav Kadavath, Tom Conerly, Amanda Askell, Tom Henighan, Dawn Drain, Ethan Perez, Nicholas Schiefer, Zac Hatfield-Dodds, Nova DasSarma, Eli Tran-Johnson, et~al.
\newblock Language models (mostly) know what they know.
\newblock \emph{arXiv preprint arXiv:2207.05221}, 2022.

\bibitem[Kaur et~al.(2024)Kaur, Samplawski, Cobb, Roy, Matejek, Acharya, Elenius, Berenbeim, Pavlik, Bastian, et~al.]{kaur2024addressing}
Ramneet Kaur, Colin Samplawski, Adam~D Cobb, Anirban Roy, Brian Matejek, Manoj Acharya, Daniel Elenius, Alexander~Michael Berenbeim, John~A Pavlik, Nathaniel~D Bastian, et~al.
\newblock Addressing uncertainty in llms to enhance reliability in generative ai.
\newblock In \emph{Neurips Safe Generative AI Workshop 2024}, 2024.

\bibitem[Kuhn et~al.(2023)Kuhn, Gal, and Farquhar]{kuhn2023semantic}
Lorenz Kuhn, Yarin Gal, and Sebastian Farquhar.
\newblock Semantic uncertainty: Linguistic invariances for uncertainty estimation in natural language generation.
\newblock In \emph{The Eleventh International Conference on Learning Representations}, 2023.

\bibitem[Kumar et~al.(2023)Kumar, Lu, Gupta, Palepu, Bellamy, Raskar, and Beam]{kumar2023conformal}
Bhawesh Kumar, Charlie Lu, Gauri Gupta, Anil Palepu, David Bellamy, Ramesh Raskar, and Andrew Beam.
\newblock Conformal prediction with large language models for multi-choice question answering.
\newblock \emph{arXiv preprint arXiv:2305.18404}, 2023.

\bibitem[Lin et~al.(2024)Lin, Trivedi, and Sun]{lingenerating}
Zhen Lin, Shubhendu Trivedi, and Jimeng Sun.
\newblock Generating with confidence: Uncertainty quantification for black-box large language models.
\newblock \emph{Transactions on Machine Learning Research}, 2024.

\bibitem[Liu et~al.(2023)Liu, Li, Wu, and Lee]{liu2023visual}
Haotian Liu, Chunyuan Li, Qingyang Wu, and Yong~Jae Lee.
\newblock Visual instruction tuning.
\newblock \emph{Advances in neural information processing systems}, 2023.

\bibitem[Liu et~al.(2024)Liu, Li, Li, and Lee]{liu2024improved}
Haotian Liu, Chunyuan Li, Yuheng Li, and Yong~Jae Lee.
\newblock Improved baselines with visual instruction tuning.
\newblock In \emph{Proceedings of the IEEE/CVF Conference on Computer Vision and Pattern Recognition}, 2024.

\bibitem[Mohri \& Hashimoto(2024)Mohri and Hashimoto]{mohri2024language}
Christopher Mohri and Tatsunori Hashimoto.
\newblock Language models with conformal factuality guarantees.
\newblock In \emph{International Conference on Machine Learning}, 2024.

\bibitem[Newcombe(1998)]{newcombe1998two}
Robert~G Newcombe.
\newblock Two-sided confidence intervals for the single proportion: comparison of seven methods.
\newblock \emph{Statistics in medicine}, 1998.

\bibitem[Ni et~al.(2025)Ni, Wang, Cheng, Blasch, and Derr]{ni2025towards}
Bo~Ni, Yu~Wang, Lu~Cheng, Erik Blasch, and Tyler Derr.
\newblock Towards trustworthy knowledge graph reasoning: An uncertainty aware perspective.
\newblock In \emph{Proceedings of the AAAI Conference on Artificial Intelligence}, 2025.

\bibitem[Nikitin et~al.(2024)Nikitin, Kossen, Gal, and Marttinen]{NEURIPS2024_10c456d2}
Alexander Nikitin, Jannik Kossen, Yarin Gal, and Pekka Marttinen.
\newblock Kernel language entropy: Fine-grained uncertainty quantification for llms from semantic similarities.
\newblock In \emph{Advances in Neural Information Processing Systems}, 2024.

\bibitem[Papadopoulos et~al.(2002)Papadopoulos, Proedrou, Vovk, and Gammerman]{papadopoulos2002inductive}
Harris Papadopoulos, Kostas Proedrou, Volodya Vovk, and Alex Gammerman.
\newblock Inductive confidence machines for regression.
\newblock In \emph{Machine learning: ECML 2002: 13th European conference on machine learning Helsinki, Finland, August 19--23, 2002 proceedings 13}, 2002.

\bibitem[Penny-Dimri et~al.(2025)Penny-Dimri, Bachmann, Cooke, Mathewlynn, Dockree, Tolladay, Kossen, Li, Gal, and Jones]{penny2025reducing}
Jahan~C Penny-Dimri, Magdalena Bachmann, William~R Cooke, Sam Mathewlynn, Samuel Dockree, John Tolladay, Jannik Kossen, Lin Li, Yarin Gal, and Gabriel~Davis Jones.
\newblock Reducing large language model safety risks in women's health using semantic entropy.
\newblock \emph{arXiv preprint arXiv:2503.00269}, 2025.

\bibitem[Quach et~al.(2024)Quach, Fisch, Schuster, Yala, Sohn, Jaakkola, and Barzilay]{quach2024conformal}
Victor Quach, Adam Fisch, Tal Schuster, Adam Yala, Jae~Ho Sohn, Tommi~S. Jaakkola, and Regina Barzilay.
\newblock Conformal language modeling.
\newblock In \emph{The Twelfth International Conference on Learning Representations}, 2024.

\bibitem[Reimers \& Gurevych(2019)Reimers and Gurevych]{reimers-gurevych-2019-sentence}
Nils Reimers and Iryna Gurevych.
\newblock Sentence-{BERT}: Sentence embeddings using {S}iamese {BERT}-networks.
\newblock In \emph{Proceedings of the 2019 Conference on Empirical Methods in Natural Language Processing and the 9th International Joint Conference on Natural Language Processing (EMNLP-IJCNLP)}, 2019.

\bibitem[Sanh et~al.(2019)Sanh, Debut, Chaumond, and Wolf]{sanh2019distilbert}
Victor Sanh, Lysandre Debut, Julien Chaumond, and Thomas Wolf.
\newblock Distilbert, a distilled version of bert: smaller, faster, cheaper and lighter.
\newblock \emph{arXiv preprint arXiv:1910.01108}, 2019.

\bibitem[Shahrokhi et~al.(2025)Shahrokhi, Roy, Yan, Arnaoudova, and Doppa]{shahrokhi2025conformal}
Hooman Shahrokhi, Devjeet~Raj Roy, Yan Yan, Venera Arnaoudova, and Janaradhan~Rao Doppa.
\newblock Conformal prediction sets for deep generative models via reduction to conformal regression.
\newblock \emph{arXiv preprint arXiv:2503.10512}, 2025.

\bibitem[Snell et~al.(2023)Snell, Zollo, Deng, Pitassi, and Zemel]{snell2023quantile}
Jake Snell, Thomas~P Zollo, Zhun Deng, Toniann Pitassi, and Richard Zemel.
\newblock Quantile risk control: A flexible framework for bounding the probability of high-loss predictions.
\newblock In \emph{The Eleventh International Conference on Learning Representations}, 2023.

\bibitem[Su et~al.(2024)Su, Wang, Ai, Hu, Wu, Zhou, and Liu]{su-etal-2024-unsupervised}
Weihang Su, Changyue Wang, Qingyao Ai, Yiran Hu, Zhijing Wu, Yujia Zhou, and Yiqun Liu.
\newblock Unsupervised real-time hallucination detection based on the internal states of large language models.
\newblock In \emph{Findings of the Association for Computational Linguistics: ACL 2024}, 2024.

\bibitem[Talmor et~al.(2019)Talmor, Herzig, Lourie, and Berant]{talmor-etal-2019-commonsenseqa}
Alon Talmor, Jonathan Herzig, Nicholas Lourie, and Jonathan Berant.
\newblock {C}ommonsense{QA}: A question answering challenge targeting commonsense knowledge.
\newblock In \emph{Proceedings of the 2019 Conference of the North {A}merican Chapter of the Association for Computational Linguistics: Human Language Technologies}, 2019.

\bibitem[Tayebati et~al.(2025)Tayebati, Kumar, Darabi, Jayasuriya, Krishnan, and Trivedi]{tayebati2025learning}
Sina Tayebati, Divake Kumar, Nastaran Darabi, Dinithi Jayasuriya, Ranganath Krishnan, and Amit~Ranjan Trivedi.
\newblock Learning conformal abstention policies for adaptive risk management in large language and vision-language models.
\newblock \emph{arXiv preprint arXiv:2502.06884}, 2025.

\bibitem[Thulin(2013)]{thulin2013cost}
M{\aa}ns Thulin.
\newblock The cost of using exact confidence intervals for a binomial proportion.
\newblock \emph{arXiv preprint arXiv:1303.1288}, 2013.

\bibitem[Vejnarov{\'a}(2000)]{vejnarova2000conditional}
Ji{\v{r}}ina Vejnarov{\'a}.
\newblock Conditional independence relations in possibility theory.
\newblock \emph{International Journal of Uncertainty, Fuzziness and Knowledge-based Systems}, 2000.

\bibitem[Wadsworth et~al.(1961)Wadsworth, Bryan, and Eringen]{wadsworth1961introduction}
George~Proctor Wadsworth, Joseph~G Bryan, and A~Cemal Eringen.
\newblock Introduction to probability and random variables.
\newblock \emph{Journal of Applied Mechanics}, 1961.

\bibitem[Wang et~al.(2024{\natexlab{a}})Wang, Bai, Tan, Wang, Fan, Bai, Chen, Liu, Wang, Ge, et~al.]{wang2024qwen2}
Peng Wang, Shuai Bai, Sinan Tan, Shijie Wang, Zhihao Fan, Jinze Bai, Keqin Chen, Xuejing Liu, Jialin Wang, Wenbin Ge, et~al.
\newblock Qwen2-vl: Enhancing vision-language model's perception of the world at any resolution.
\newblock \emph{arXiv preprint arXiv:2409.12191}, 2024{\natexlab{a}}.

\bibitem[Wang et~al.(2025{\natexlab{a}})Wang, Geng, Wang, Wang, Fu, and Zheng]{wang2025sample}
Qingni Wang, Tiantian Geng, Zhiyuan Wang, Teng Wang, Bo~Fu, and Feng Zheng.
\newblock Sample then identify: A general framework for risk control and assessment in multimodal large language models.
\newblock In \emph{The Thirteenth International Conference on Learning Representations}, 2025{\natexlab{a}}.

\bibitem[Wang et~al.(2024{\natexlab{b}})Wang, Duan, Cheng, Zhang, Wang, Shi, Xu, Shen, and Zhu]{wang-etal-2024-conu}
Zhiyuan Wang, Jinhao Duan, Lu~Cheng, Yue Zhang, Qingni Wang, Xiaoshuang Shi, Kaidi Xu, Heng~Tao Shen, and Xiaofeng Zhu.
\newblock {C}on{U}: Conformal uncertainty in large language models with correctness coverage guarantees.
\newblock In \emph{Findings of the Association for Computational Linguistics: EMNLP 2024}, 2024{\natexlab{b}}.

\bibitem[Wang et~al.(2025{\natexlab{b}})Wang, Duan, Yuan, Chen, Chen, Zhang, Wang, Shi, and Xu]{wang2025word}
Zhiyuan Wang, Jinhao Duan, Chenxi Yuan, Qingyu Chen, Tianlong Chen, Yue Zhang, Ren Wang, Xiaoshuang Shi, and Kaidi Xu.
\newblock Word-sequence entropy: Towards uncertainty estimation in free-form medical question answering applications and beyond.
\newblock \emph{Engineering Applications of Artificial Intelligence}, 2025{\natexlab{b}}.

\bibitem[Wang et~al.(2025{\natexlab{c}})Wang, Wang, Zhang, Chen, Zhu, Shi, and Xu]{wang2025sconu}
Zhiyuan Wang, Qingni Wang, Yue Zhang, Tianlong Chen, Xiaofeng Zhu, Xiaoshuang Shi, and Kaidi Xu.
\newblock Sconu: Selective conformal uncertainty in large language models.
\newblock \emph{arXiv preprint arXiv:2504.14154}, 2025{\natexlab{c}}.

\bibitem[Yadkori et~al.(2024)Yadkori, Kuzborskij, Stutz, Gy{\"o}rgy, Fisch, Doucet, Beloshapka, Weng, Yang, Szepesv{\'a}ri, et~al.]{yadkori2024mitigating}
Yasin~Abbasi Yadkori, Ilja Kuzborskij, David Stutz, Andr{\'a}s Gy{\"o}rgy, Adam Fisch, Arnaud Doucet, Iuliya Beloshapka, Wei-Hung Weng, Yao-Yuan Yang, Csaba Szepesv{\'a}ri, et~al.
\newblock Mitigating llm hallucinations via conformal abstention.
\newblock \emph{arXiv preprint arXiv:2405.01563}, 2024.

\bibitem[Yang et~al.(2024)Yang, Yang, Hui, Zheng, Yu, Zhou, Li, Li, Liu, Huang, Dong, Wei, Lin, Tang, Wang, Yang, Tu, Zhang, Ma, Xu, Zhou, Bai, He, Lin, Dang, Lu, Chen, Yang, Li, Xue, Ni, Zhang, Wang, Peng, Men, Gao, Lin, Wang, Bai, Tan, Zhu, Li, Liu, Ge, Deng, Zhou, Ren, Zhang, Wei, Ren, Fan, Yao, Zhang, Wan, Chu, Cui, Zhang, and Fan]{Yang2024Qwen2TR}
An~Yang, Baosong Yang, Binyuan Hui, Bo~Zheng, Bowen Yu, Chang Zhou, Chengpeng Li, Chengyuan Li, Dayiheng Liu, Fei Huang, Guanting Dong, Haoran Wei, Huan Lin, Jialong Tang, Jialin Wang, Jian Yang, Jianhong Tu, Jianwei Zhang, Jianxin Ma, Jin Xu, Jingren Zhou, Jinze Bai, Jinzheng He, Junyang Lin, Kai Dang, Keming Lu, Ke-Yang Chen, Kexin Yang, Mei Li, Min Xue, Na~Ni, Pei Zhang, Peng Wang, Ru~Peng, Rui Men, Ruize Gao, Runji Lin, Shijie Wang, Shuai Bai, Sinan Tan, Tianhang Zhu, Tianhao Li, Tianyu Liu, Wenbin Ge, Xiaodong Deng, Xiaohuan Zhou, Xingzhang Ren, Xinyu Zhang, Xipin Wei, Xuancheng Ren, Yang Fan, Yang Yao, Yichang Zhang, Yunyang Wan, Yunfei Chu, Zeyu Cui, Zhenru Zhang, and Zhi-Wei Fan.
\newblock Qwen2 technical report.
\newblock \emph{arXiv preprint arXiv:2407.10671}, 2024.

\bibitem[Yang et~al.(2025)Yang, Zhang, Zhang, Huang, Yu, Collier, and Yang]{yang2025uncle}
Ruihan Yang, Caiqi Zhang, Zhisong Zhang, Xinting Huang, Dong Yu, Nigel Collier, and Deqing Yang.
\newblock Uncle: Uncertainty expressions in long-form generation.
\newblock \emph{arXiv preprint arXiv:2505.16922}, 2025.

\bibitem[Yao et~al.(2024)Yao, Duan, Xu, Cai, Sun, and Zhang]{yao2024survey}
Yifan Yao, Jinhao Duan, Kaidi Xu, Yuanfang Cai, Zhibo Sun, and Yue Zhang.
\newblock A survey on large language model (llm) security and privacy: The good, the bad, and the ugly.
\newblock \emph{High-Confidence Computing}, 2024.

\bibitem[Ye et~al.(2024)Ye, Yang, Pang, Wang, Wong, Yilmaz, Shi, and Tu]{ye2024benchmarking}
Fanghua Ye, Mingming Yang, Jianhui Pang, Longyue Wang, Derek Wong, Emine Yilmaz, Shuming Shi, and Zhaopeng Tu.
\newblock Benchmarking llms via uncertainty quantification.
\newblock \emph{Advances in Neural Information Processing Systems}, 2024.

\bibitem[Yu et~al.(2024)Yu, Yang, Li, Wang, Lin, Liu, Wang, and Wang]{yu2024mmvet}
Weihao Yu, Zhengyuan Yang, Linjie Li, Jianfeng Wang, Kevin Lin, Zicheng Liu, Xinchao Wang, and Lijuan Wang.
\newblock {MM}-vet: Evaluating large multimodal models for integrated capabilities.
\newblock In \emph{Forty-first International Conference on Machine Learning}, 2024.

\bibitem[Zhang et~al.(2024)Zhang, Zhang, and Zheng]{zhang2024vl}
Ruiyang Zhang, Hu~Zhang, and Zhedong Zheng.
\newblock Vl-uncertainty: Detecting hallucination in large vision-language model via uncertainty estimation.
\newblock \emph{arXiv preprint arXiv:2411.11919}, 2024.

\bibitem[Zhao et~al.(2025)Zhao, Yuan, Yang, and Naseem]{zhao2025can}
Shangziqi Zhao, Jiahao Yuan, Guisong Yang, and Usman Naseem.
\newblock Can pruning improve reasoning? revisiting long-cot compression with capability in mind for better reasoning.
\newblock \emph{arXiv preprint arXiv:2505.14582}, 2025.

\bibitem[Zheng et~al.(2023)Zheng, Chiang, Sheng, Zhuang, Wu, Zhuang, Lin, Li, Li, Xing, Zhang, Gonzalez, and Stoica]{vicuna2023}
Lianmin Zheng, Wei-Lin Chiang, Ying Sheng, Siyuan Zhuang, Zhanghao Wu, Yonghao Zhuang, Zi~Lin, Zhuohan Li, Dacheng Li, Eric~P. Xing, Hao Zhang, Joseph~E. Gonzalez, and Ion Stoica.
\newblock Judging llm-as-a-judge with mt-bench and chatbot arena.
\newblock In \emph{Thirty-seventh Conference on Neural Information Processing Systems}, 2023.

\bibitem[Zheng et~al.(2025)Zheng, Gan, Chen, Qi, Liang, and Yu]{zheng2025large}
Yanxin Zheng, Wensheng Gan, Zefeng Chen, Zhenlian Qi, Qian Liang, and Philip~S Yu.
\newblock Large language models for medicine: a survey.
\newblock \emph{International Journal of Machine Learning and Cybernetics}, 2025.

\bibitem[Zollo et~al.(2024)Zollo, Morrill, Deng, Snell, Pitassi, and Zemel]{zollo2024prompt}
Thomas~P Zollo, Todd Morrill, Zhun Deng, Jake Snell, Toniann Pitassi, and Richard Zemel.
\newblock Prompt risk control: A rigorous framework for responsible deployment of large language models.
\newblock In \emph{The Twelfth International Conference on Learning Representations}, 2024.

\end{thebibliography}
\bibliographystyle{iclr2025_conference}

\newpage

\appendix
\section{Bernoulli Structure under Conditional Selection to I.I.D. QA Samples}\label{appendix: statistical Structure}
We first justify the i.i.d. property of the subset $\mathcal{S}_t$. 
Recall that the original QA data points $\left(X_i, Y_i^*\right)$ are i.i.d. according to the joint distribution $\mathcal{D}$, i.e., $\left(X_i, Y_i^*\right) \mathop{\sim}\limits^{i.i.d.} \mathcal{D}, \forall i$. 
Once the uncertainty threshold is set to $t$, the selection indicator $I_i := \mathbf{1}\left\{U(F(X_i)) \le t\right\}$ depends solely on the model's uncertainty to $X_i$, and $U(F(X_i))$ is a deterministic function of $X_i$. 
We redefine the selected subset $\mathcal{S}_t$ drawn from the conditional distribution $\mathcal{D}_t$ as
\[\left(X_i, Y_i^*\right)\mid I_i=1 \sim \mathcal{D}_t=\mathcal{D}\mid U(F(X)) \le t,\]
where the selection indicator $I_i=1$ is equivalent to the event $U(F(X_i)) \le t$, and the conditional distribution of selected samples is the restriction of $\mathcal{D}$ to the event $U(F(X)) \le t$. 
By definition, all selected samples satisfy $U(F(X_i)) \le t$, so they are drawn from the same conditional distribution $\mathcal{D}_t$. 
For example, if the original data follows a normal distribution $\mathcal{N}\left(0,1\right)$, the selected subset’s distribution would be a truncated version of $\mathcal{N}\left(0,1\right)$ within the interval defined by $t$. 
Therefore, for any QA data points $\left(X_i, Y_i^*\right)\in \mathcal{S}_t$, its distribution satisfies $\left(X_i, Y_i^*\right) \sim \mathcal{D}_t$. 

Since the original data is independent, the joint probability before selection can be decomposed as 
\begin{equation}
    \mathrm{Pr}\left( \displaystyle\bigcap_{k=1}^{K} \left(X_{i_k}, Y_{i_k}^*\right)\in B_k \right) = \displaystyle\prod_{k=1}^{K}  \mathrm{Pr} \left( \left(X_{i_k}, Y_{i_k}^*\right)\in B_k \right),
\end{equation}
where $\left\{ \left(X_{i_k}, Y_{i_k}^*\right) \right\}_{k=1}^{K}$ ($\subseteq \mathcal{S}_t$) is any finite collection of selected data points, and $B_k$ are measurable sets~\cite{vejnarova2000conditional,batu2001testing}. 
Conditioning on the selection events $I_{i_k}=1$, the joint probability becomes $\mathrm{Pr}\left( \textstyle\bigcap_{k=1}^{K} \left(X_{i_k}, Y_{i_k}^*\right)\in B_k \mid I_{i_k} =1 \right)$. 
Since the selection indicators $I_{i_k}=1$ depend only on $X_{i_k}$, and the original QA data points are independent, the selection events $I_{i_k}=1$ are mutually independent. 
Thus, the joint conditional probability can be factorized as $\textstyle\prod_{k=1}^{K}  \mathrm{Pr} \left( \left(X_{i_k}, Y_{i_k}^*\right)\in B_k \mid I_{i_k}=1 \right)$. 
By the definition of the conditional distribution $\mathcal{D}_t$, each conditional probability is equivalent to $\mathrm{Pr} \left( \left(X,Y^*\right)\in B_k \mid U(F(X)) \leq t \right)$. 
Therefore, for any finite collection $\left\{ \left(X_{i_k}, Y_{i_k}^*\right) \right\}_{k=1}^{K} \subseteq \mathcal{S}_t$, we have 
\begin{equation}
    \mathrm{Pr}\left( \displaystyle\bigcap_{k=1}^{K} \left(X_{i_k}, Y_{i_k}^*\right)\in B_k \mid I_{i_k} =1 \right) = \displaystyle\prod_{k=1}^{K}  \mathrm{Pr} \left( \left(X_{i_k}, Y_{i_k}^*\right)\in B_k \mid U(F(X)) \leq t \right).
\end{equation}
We conclude that, under the uncertainty-guarding selection mechanism, the subset $\mathcal{S}_t$ retains the i.i.d. property of original samples, i.e., $\left(X_i, Y_i^*\right) \mathop{\sim}\limits^{i.i.d.} \mathcal{D}_t$, where independence holds because the joint probability equals the product of marginal probabilities, and identical distribution holds because each probability term originates from the same conditional distribution $\mathcal{D}_t$. 

We then justify the Bernoulli property of $Z_i$. 
Since the ground truth $Y_i^*$ is generated independently of the model output $\hat{Y}$, $Z_i = \mathbf{1} \left\{ A\left( \hat{Y}, Y_i^* \right)=1 \right\}$ depend on $Y_i^* \mid X_i$, which is independent of $\hat{Y}$. 
Given that the selection indicator $I_i$ depends only on $X_i$, we have $Z_i \perp I_i \mid X_i$. 
That is,
\begin{equation}
    \mathrm{Pr}\left(Z_i = z \mid I_i=1, X_i\right) = \mathrm{Pr}\left(Z_i=z\right).
\end{equation}
For each selected sample $\left(X_i, Y_i^*\right) \in \mathcal{S}_t$, the correctness $Z_i$ is determined by $Y_i^* \mid X_i$, which is independent across $i$ due to the i.i.d. property of QA data points. 
Finally, the success probability is homogeneous across samples: 
\begin{equation}
\begin{split}
    \mathrm{Pr}\left(Z_i = z \mid I_i=1\right) &= \mathbb{E}_{(X, Y^*)\sim \mathcal{D}_t} \left[ \mathbf{1}\left\{A\left( \hat{Y}, Y^* \right)=1\right\} \right]\\
    &=1-R(t). 
\end{split}
\end{equation}
Thus, $\left\{Z_i\right\}_{(X, Y^*)\in \mathcal{S}_t} \mathop{\sim}\limits^{i.i.d.} \mathrm{Bernoulli} \left(1-R(t)\right)$. 
Equivalently, the failure indicators $W_i := 1 - Z_i \sim \mathrm{Bernoulli} \left(R\left(t\right) \right)$ are also i.i.d. variables under this conditional distribution $\mathcal{D}_t$. 


\section{Proofs}\label{appendix: proofs}
\paragraph{Clopper–Pearson Exact Upper Confidence Bound.} 
In this section, we first provide a complete proof that the upper confidence bound $\hat{R}^{\text{upper}}_{1-\delta}(t)$ defined in Eq.~\ref{eq: upper bound via CDF} satisfies $\mathrm{Pr}\left( R\left(t\right) \leq \hat{R}_{1-\delta}^{\text{upper}}\left(t\right) \right) \geq 1 - \delta$ defined in Eq.~\ref{eq: probability inequality of upper bound}. 

Recall the CDF \( D(r \mid R(t)) \) of \( \hat{R}(t) \) under the true failure rate \( R(t) \) in Eq.~\ref{eq: cdf}, we then define the corresponding inverse CDF as 
\begin{equation}
    D^{-1}\left( p \right) = \sup \left\{  r: D\left(r \mid  R\left( t \right)\right) \leq p \right\}
\end{equation}

By the definition of $\hat{R}_{1-\delta}^{\text{upper}}\left(t\right)$, it holds that 
\begin{equation}
    D\left( \hat{r}_{cal}\left(t\right) \mid \hat{R}_{1-\delta}^{\text{upper}}\left(t\right) \right)  = \delta.
\end{equation}
Since $D^{-1}\left(\delta\right) = \sup \left\{  r: D\left(r \mid  R\left( t \right)\right) \leq \delta \right\}$, if $D\left(\hat{r}_{cal}\left(t\right)\mid  R\left( t \right)\right) < \delta$, then we have $\hat{r}_{cal}\left(t\right) < D^{-1}\left(\delta\right)$. 
This implies the following equivalence
\[
\begin{split}
    &\left\{D\left(\hat{r}_{cal}\left(t\right)\mid  R\left( t \right)\right) < \delta \Rightarrow \hat{r}_{cal}\left(t\right) < D^{-1}\left(\delta\right)\right\}\\& \Longleftrightarrow   \left\{ \mathrm{Pr} \left( D\left(\hat{r}_{cal}\left(t\right) \mid R\left( t \right)\right) < \delta \right) \leq \mathrm{Pr} \left( \hat{r}_{cal}\left(t\right) < D^{-1}\left(\delta\right)\right) \right\} 
\end{split}
\]

Considering that $D\left(r \mid R\left( t \right) \right)$ is monotonic decreasing in $R\left(t\right)$, i.e., larger $R\left(t\right)$ leads to smaller $D\left(r \mid R\left( t \right) \right)$, if $R\left( t \right) > \hat{R}_{1-\delta}^{\text{upper}}\left(t\right)$, then $D\left(\hat{r}_{cal}\left(t\right) \mid R\left( t \right) \right) < \delta$. 
This implies 
\[
\begin{split}
    &\left\{R\left( t \right) > \hat{R}_{1-\delta}^{\text{upper}}\left(t\right) \Rightarrow D\left(\hat{r}_{cal}\left(t\right) \mid R\left( t \right) \right) < \delta\right\}\\& \Longleftrightarrow   \left\{ \mathrm{Pr}\left(R\left( t \right) > \hat{R}_{1-\delta}^{\text{upper}}\left(t\right)\right) \leq \mathrm{Pr} \left( D\left(\hat{r}_{cal}\left(t\right) \mid R\left( t \right)\right) < \delta \right) \right\} 
\end{split}
\]

Thus,
\begin{equation}
\begin{split}
    \mathrm{Pr}\left(R\left( t \right) \leq \hat{R}_{1-\delta}^{\text{upper}}\left(t\right)\right) &= 1 -  \mathrm{Pr}\left(R\left( t \right) > \hat{R}_{1-\delta}^{\text{upper}}\left(t\right)\right) \\
    & \geq 1 - \mathrm{Pr} \left( D\left(\hat{r}_{cal}\left(t\right) \mid R\left( t \right)\right) < \delta \right)\\
    & \geq 1 - \mathrm{Pr} \left( \hat{r}_{cal}\left(t\right) < D^{-1}\left(\delta\right)\right)\\
    & \geq 1-\delta
\end{split},
\end{equation}
demonstrating that $\hat{R}_{1-\delta}^{\text{upper}}\left(t\right)$ is the upper endpoint of the Clopper–Pearson exact $1-\delta$ confidence interval for the TCFR $R\left( t \right)$. 

\paragraph{Hoeffding-style Upper Confidence Bound.} 
We then provide a formal proof of Lemma~\ref{lm: Hoeffding-style Upper Confidence Bound} to justify the statistical validity of $\hat{R}^{\text{upper-H}}_{1-\delta}(t)$ defined in Eq.~\ref{eq: upper-h}. 

We begin from the general form of Hoeffding’s inequality~\cite{hoeffding1994probability}. 
Let \( W_1, \ldots, W_m \in \{0,1\} \) be independent random variables with $W_i\in \left[a_i, b_i\right]$, and let $\mu = \mathbb{E}\left(W_i\right)$. 
Then for all $t \geq 0$, 
\begin{equation}\label{eq: general Hoeffding’s inequality}
    \mathrm{Pr} \left( \left| \frac{1}{m} \sum_{i=1}^{m} W_i - \mu \right| \geq t \right) \leq 2 \exp \left(-\frac{2m^2t^2}{\sum_{i=1}^{m} \left(b_i-a_i\right)^2}\right).
\end{equation}
In our settings, each $W_i \in \left\{0, 1\right\}$, so $a_i=0, b_i=1$, and $b_i-a_i=1$ for all $i$. 
Thus, we obtain
\begin{equation}\label{eq: b-a=m}
    \sum_{i=1}^{m} \left(b_i-a_i\right)^2=m.
\end{equation}
Plugging Eq.~\ref{eq: b-a=m} into Eq.~\ref{eq: general Hoeffding’s inequality}, the two-sided bound in Eq.~\ref{eq: brv Hoeffding’s inequality} holds. 

To derive the one-sided upper bound, we remove the absolute value symbol in Eq.~\ref{eq: brv Hoeffding’s inequality} and obtain
\begin{equation}\label{eq: one-side brv hb}
    \mathrm{Pr} \left( - \frac{1}{m} \sum_{i=1}^{m} W_i + \mu \geq t \right) \leq  \exp \left(-2mt^2\right).
\end{equation}
We set the right-hand side equal to $\delta$ (i.e., $\delta=\exp \left(-2mt^2\right)$) and have 
\begin{equation}\label{eq: t}
    t = \sqrt{\frac{1}{2m} \log \frac{1}{\delta}}.
\end{equation}
Hence, plugging Eq.~\ref{eq: t} and $\delta$ into Eq.~\ref{eq: one-side brv hb}, we conclude
\begin{equation}
    \mathrm{Pr} \left(  \mu \geq \frac{1}{m} \sum_{i=1}^{m} W_i + \sqrt{\frac{1}{2m} \log \frac{1}{\delta}} \right) \leq  \delta. 
\end{equation}
Equivalently, Eq.~\ref{eq: lemma 2 one-sided bound} in Lemma~\ref{lm: Hoeffding-style Upper Confidence Bound} holds with probability at least $1-\delta$. 
This completes the proof.

\section{Details of Experimental Settings}\label{appendix: Details of Experimental Settings}
\subsection{Details of Datasets}\label{appendix: Details of Datasets}
\textbf{CommonsenseQA}~\cite{talmor-etal-2019-commonsenseqa} is a multiple-choice QA dataset containing commonsense-knowledge questions, each with one correct answer and four distractor answers (i.e., five choices). 
We utilize both the full training (9,741 samples) and validation (1,221 samples) splits. 
Note that some samples contain non-ASCII characters that the tokenizer cannot encode, so we discard those samples. 
In addition, we employ one QA sample from the remaining data as a fixed example and add it as a one-shot prompt before the question of other samples. 
An example of the complete prompt is presented in Figure~\ref{fig: CommonsenseQA prompt}. 
In total, we utilize 10,956 QA samples. 

\textbf{TriviaQA}~\cite{joshi-etal-2017-triviaqa} is a high-quality closed-book dataset that contains over 650k question-answer pairs. 
Following prior work~\cite{duan2024shifting}, we randomly select 2,000 QA data points from the validation split of the \texttt{rc.nocontext} subset. 
We also develop a one-shot prompt for each data. 
An example of the complete prompt is presented in Figure~\ref{fig: TriviaQA prompt}. 

\textbf{MMVet}~\cite{yu2024mmvet} is an evaluation benchmark designed to assess LVLMs 
on complex multimodal tasks containing 218 free-form QA data points. 
It systematically structures these tasks by defining six core vision-language capabilities and evaluating 16 combinations of these capabilities to capture models' ability to integrate different skills. 
Following~\cite{zhang2024vl}, an example of the complete prompt is presented in Figure~\ref{fig: MMVet prompt}.

\subsection{Details of Base Models}\label{appendix: Details of Models}
We consider three series of open-source LLMs available: LLaMA~\citep{llama3modelcard}, Vicuna~\citep{vicuna2023}, and Qwen~\citep{Yang2024Qwen2TR}, divided by the model size into: \ding{172} 3B: Qwen-2.5-3B-Instruct. \ding{173} 7B: Qwen-2.5-7B-Instruct. \ding{174} 8B: LLaMA-3.1-8B-Instruct. \ding{175} 13B: Vicuna-13B-v1.5. \ding{176} 14B: Qwen-2.5-14B-Instruct. 
Following VL-Uncertainty~\cite{zhang2024vl}, we select five popular LVLMs from four distinct model groups: LLaVA1.5~\cite{liu2023visual}, LLaVA-NeXT~\cite{liu2024improved}, Qwen2VL~\cite{wang2024qwen2}, and InternVL2~\cite{chen2024internvl}, divided by the model size into: \ding{172} 2B: Qwen2-VL-2B-Instruct. \ding{173} 7B: Qwen2-VL-7B-Instruct. \ding{174} 8B: InternVL2-8B. \ding{175}: LLaVA-15-13B-HF and LLaVA-V1.6-Vicuna-13B-HF. 

\subsection{Details of Uncertainty Measures}\label{appendix: Details of Uncertainty Measures}
For closed-ended QA tasks, the \textit{PE} is directly computed from the option probabilities assigned by the LLM.
In the white-box setting, the softmax-transformed internal logits are used~\cite{ye2024benchmarking}.
In the black-box setting, 20 responses are sampled via multinomial sampling to estimate option-wise frequencies~\cite{wang2025sample,wang2025sconu}. 
For open-domain tasks, we utilize \textit{SE} at the black-box setting by default~\cite{farquhar2024detecting,zhang2024vl}, where we sample 10 responses per question to perform semantic clustering, and compute the \textit{SE} based on the cluster probabilities (cluster size divided by the sampling size). 
In the white-box setting, the probability of each cluster is computed as the sum of the generative probabilities based on sequence logits for all responses within the cluster~\cite{kuhn2023semantic}. 
For \textit{Ecc}, \textit{Deg}, and \textit{Eigv}, we sample 10 responses for each sample to compute the similarity matrix; the detailed procedure follows~\cite{lingenerating}.

\begin{figure*}[!t]
    \centering
    \begin{subfigure}{0.495\linewidth}
        \centering
        \includegraphics[width=\linewidth]{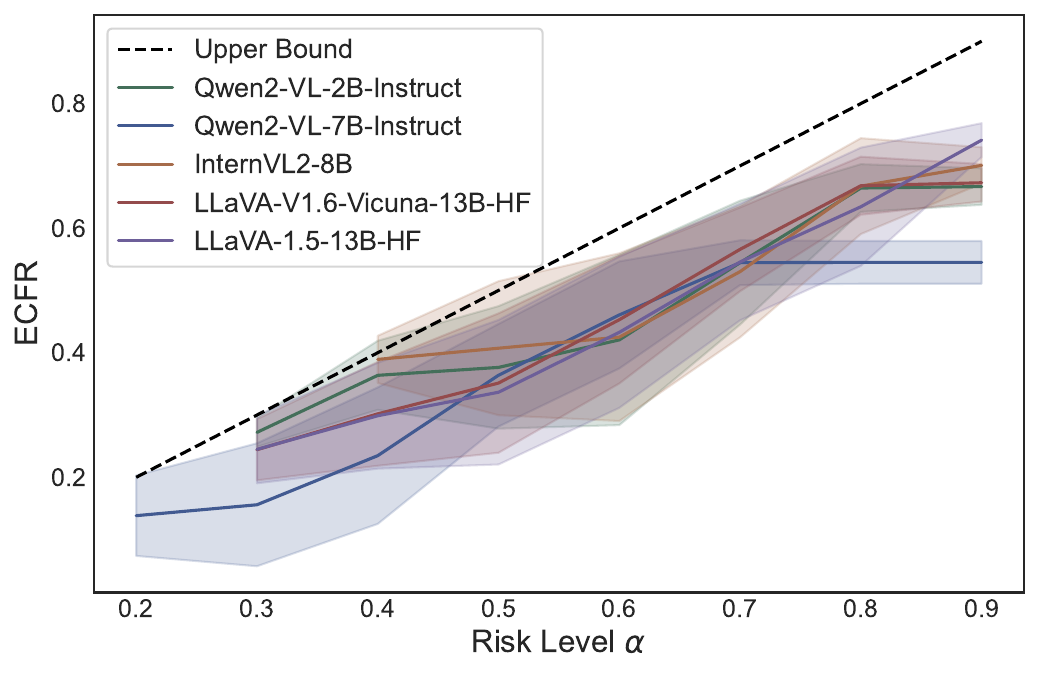}
        \caption{COIN-CP.}
	\label{fig: mmvet risk control cp black}
    \end{subfigure}
    \hfill
    \centering
    \begin{subfigure}{0.495\linewidth}
	\centering
	\includegraphics[width=\linewidth]{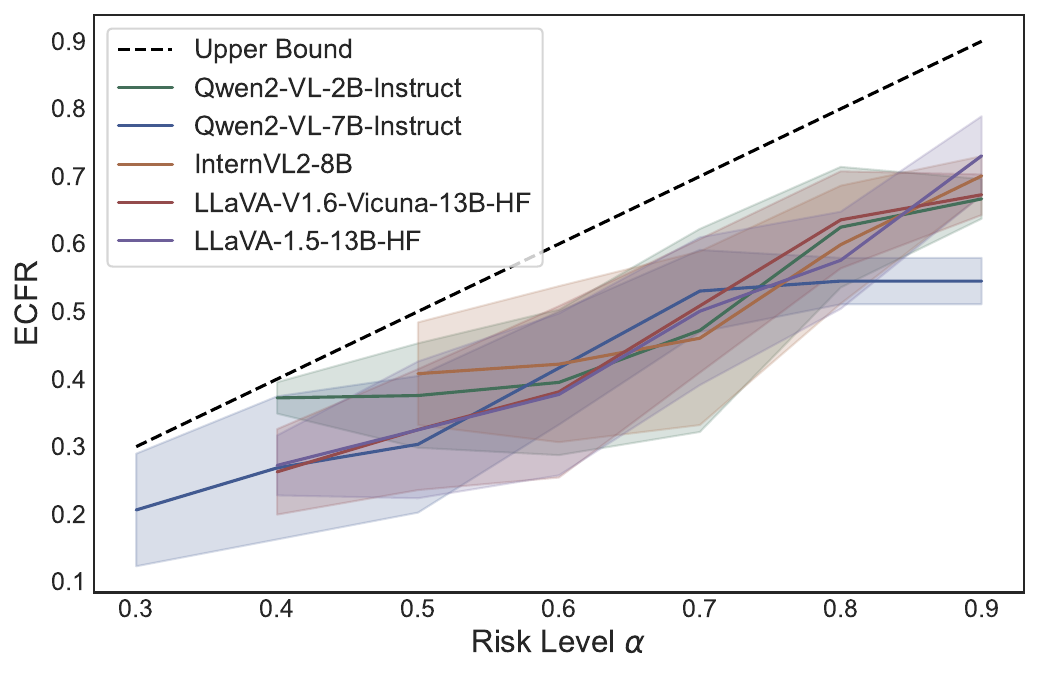}
	\caption{COIN-HFD.}
	\label{fig: mmvet risk control hfd black}
    \end{subfigure}
    \caption{Results of the ECFR over test samples on the MMVet dataset utilizing 5 LVLMs of sizes ranging from 2B to 13B.}\label{fig: mmvet results}
\end{figure*}

\begin{figure*}[!t]
    \centering
    \begin{subfigure}{0.495\linewidth}
        \centering
        \includegraphics[width=\linewidth]{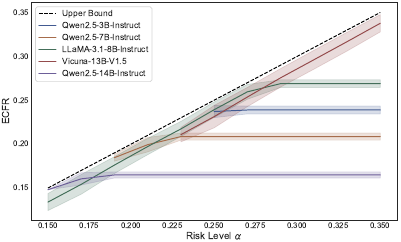}
        \caption{CommonsenseQA.}
	\label{fig: commonsenseqa risk control cp black}
    \end{subfigure}
    \hfill
    \centering
    \begin{subfigure}{0.495\linewidth}
        \centering
        \includegraphics[width=\linewidth]{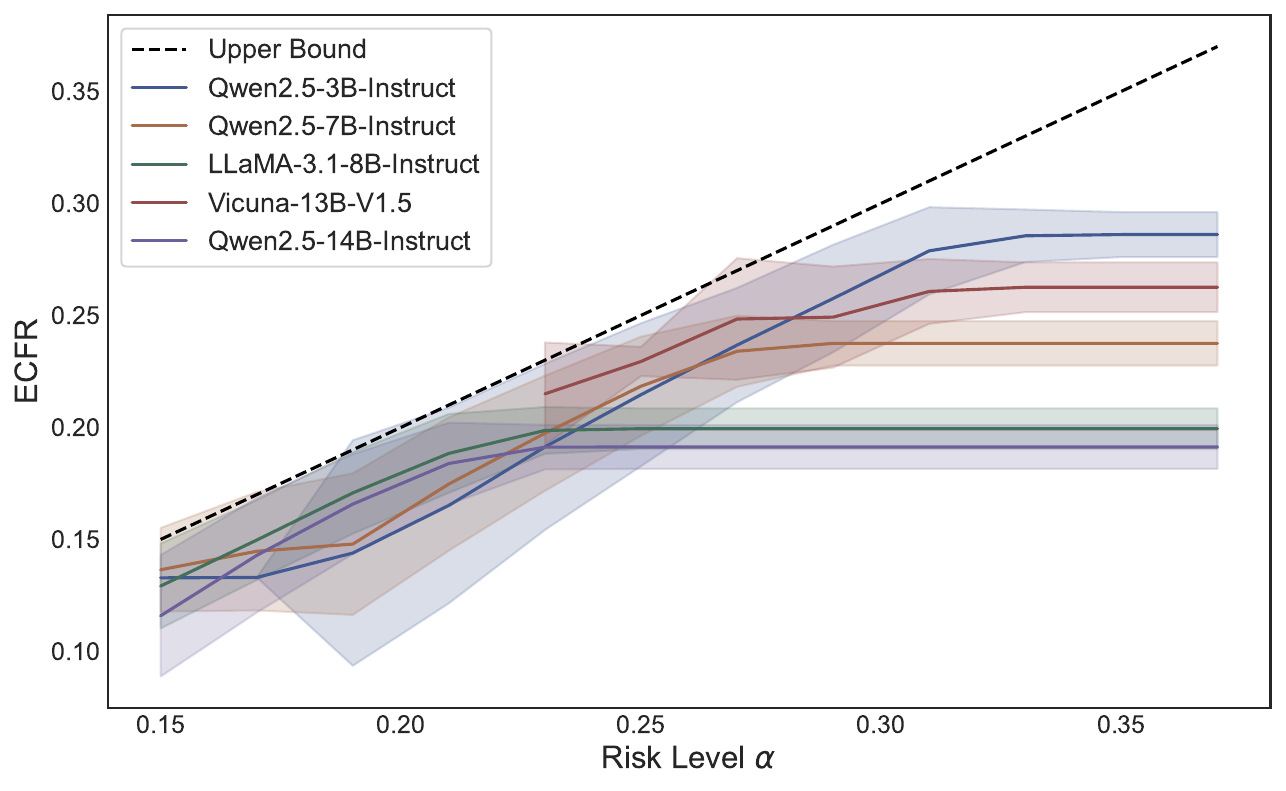}
        \caption{TriviaQA.}
	\label{fig: triviaqa risk control cp white}
    \end{subfigure}
    \caption{(a) Results of the ECFR on the CommonsenseQA dataset under the black-box setting of \textit{PE}. (b) Results of ECFR on the TriviaQA dataset under the white-box setting of \textit{SE}.}\label{fig: white-black risk control}
\end{figure*}
\newpage

\subsection{Details of Additional Admission Functions}\label{appendix: Details of Correctness Metrics} 
In addition to sentence similarity, we adopt a bidirectional entailment approach to evaluate the correctness of free-form answers~\cite{kuhn2023semantic,farquhar2024detecting,lingenerating,wang2025word}.
Specifically, we employ an off-the-shelf DeBERTa-large model~\cite{he2021deberta} as the Natural Language Inference (NLI) classifier, which outputs logits over three semantic relation classes: entailment, neutral, and contradiction.
An answer is deemed correct if the classifier predicts entailment for both directions, i.e., when evaluated on (answer $\rightarrow$ ground-truth) and (ground-truth $\rightarrow$ answer).
For MMVet, following~\cite{zhang2024vl}, we employ Qwen-2.5-3B-Instruct to assess whether answer matches the ground-truth, using the prompt template: 
\begin{tcolorbox}
Ground truth: $<$sentence 1$>$.\\
Model answer: $<$sentence 2$>$.\\
Please verify if the model answer matches the ground truth. Respond with either `Correct' or `Wrong' only.
\end{tcolorbox}


\section{Additional Experimental Results}\label{appendix: Additional Experimental Results}
\noindent\textbf{Additional Results on the MMVet Dataset.} 
We further assess the statistical validity of COIN on the multimodal MMVet dataset utilizing five representative LVLMs. 
As demonstrated in Figure~\ref{fig: mmvet results}, both COIN-CP and COIN-HDF can consistently constrain the ECFR on the test set below the upper bound across a range of user-specified risk levels. 
Due to the relatively small number of multimodal QA samples, the results of ECFR exhibits higher variance over 100 trials; nevertheless, the mean remains below the target risk level. 
As discussed in previous work~\cite{angelopoulos2021gentle,ye2024benchmarking}, occasional variance exceeding the upper bound arises from finite-sample variability—while the theoretical guarantee is rigorous, minor deviations are expected in practice.

\begin{figure*}[!t]
    \centering
    \begin{subfigure}{0.495\linewidth}
        \centering
        \includegraphics[width=\linewidth]{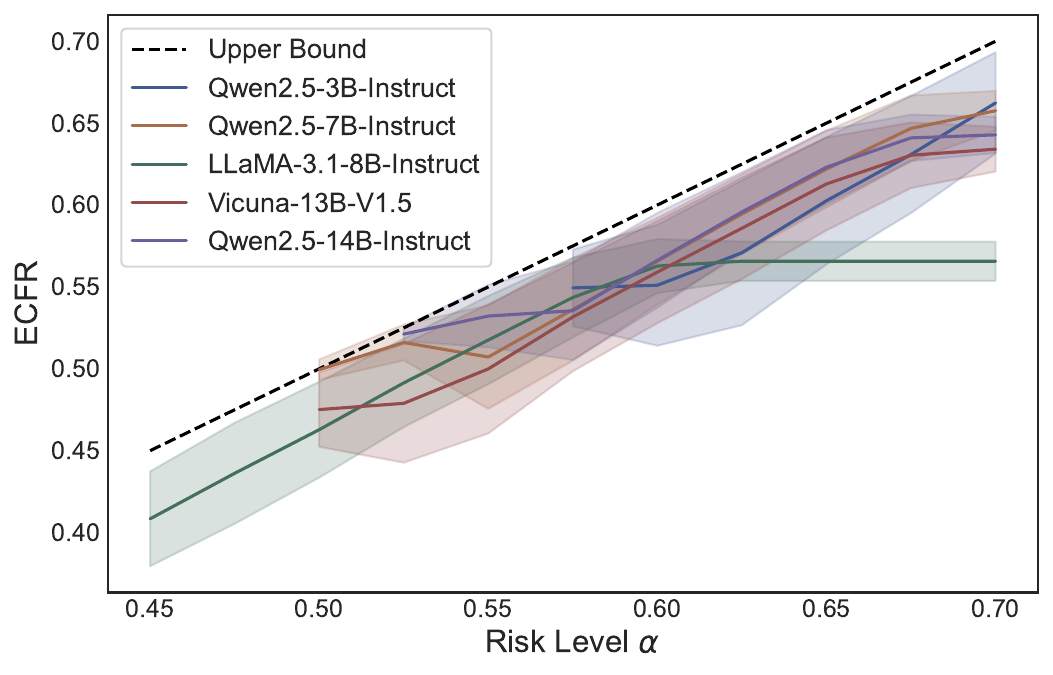}
        \caption{COIN-CP.}
	\label{fig: triviaqa risk control cp entailment}
    \end{subfigure}
    \hfill
    \centering
    \begin{subfigure}{0.495\linewidth}
        \centering
        \includegraphics[width=\linewidth]{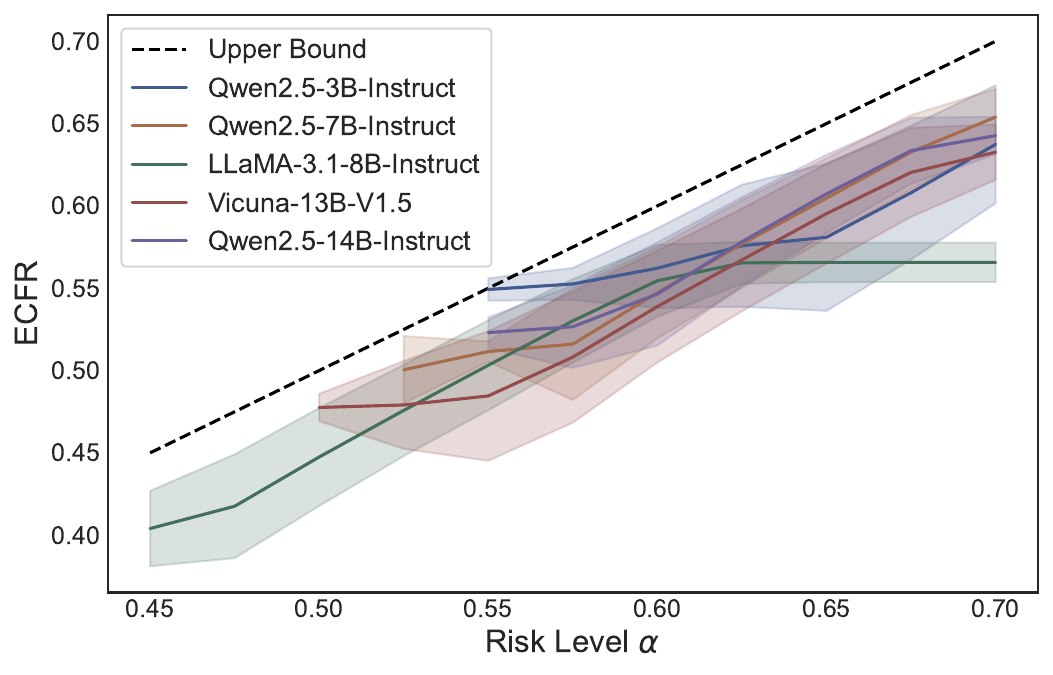}
        \caption{COIN-HFD.}
	\label{fig: triviaqa risk control hfd entailment}
    \end{subfigure}
    \caption{Results of the ECFR on the TriviaQA dataset with bi-entailment as the correctness metric. The calibration-to-test data ratio is set to 0.5.}\label{fig: triviaqa risk control entailment}
\end{figure*}

\begin{figure*}[!t]
    \centering
    \begin{subfigure}{0.495\linewidth}
        \centering
        \includegraphics[width=\linewidth]{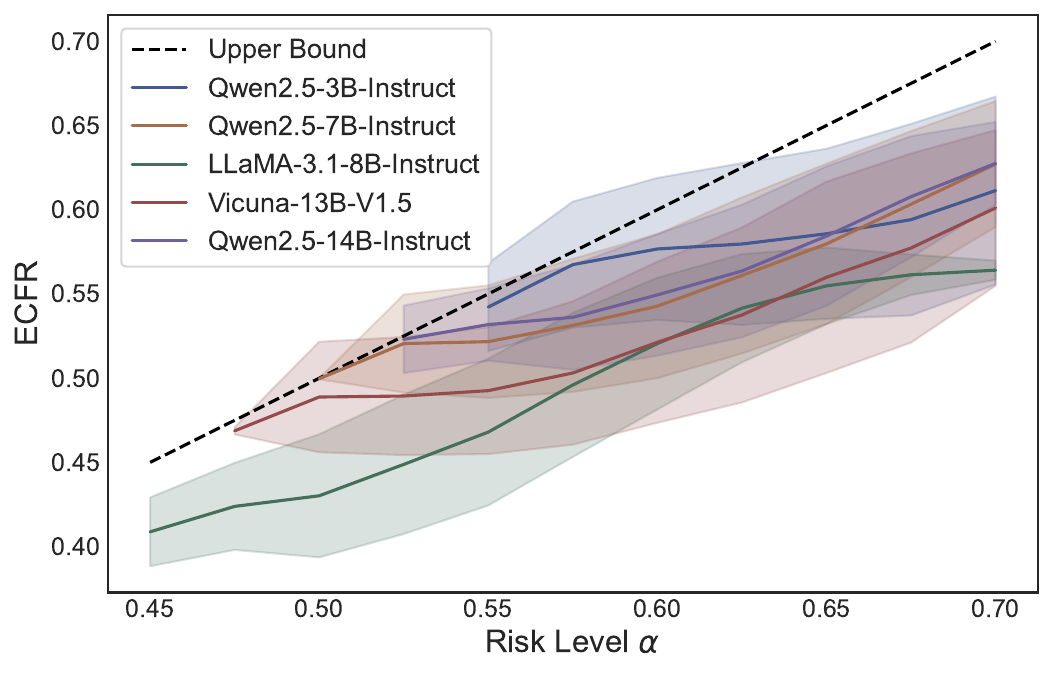}
        \caption{COIN-CP.}
	\label{fig: triviaqa risk control cp entailment 1:9}
    \end{subfigure}
    \hfill
    \centering
    \begin{subfigure}{0.495\linewidth}
        \centering
        \includegraphics[width=\linewidth]{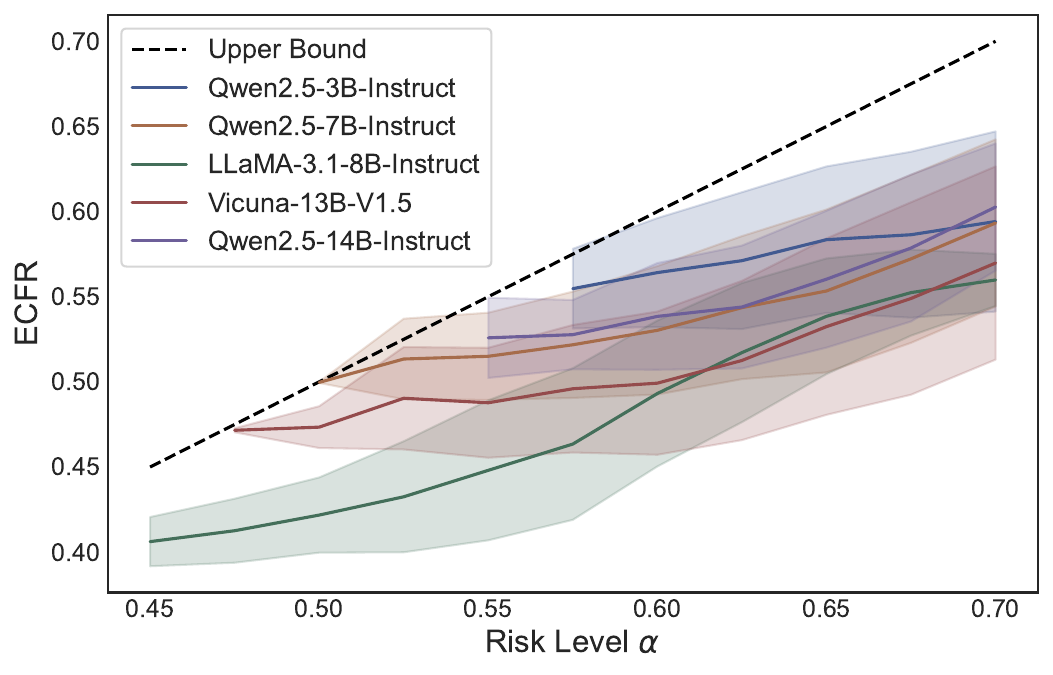}
        \caption{COIN-HFD.}
	\label{fig: triviaqa risk control hfd entailment 1:9}
    \end{subfigure}
    \caption{Results of the ECFR on the TriviaQA dataset with bi-entailment as the correctness metric. The calibration-to-test data ratio is set to 0.1.}\label{fig: triviaqa risk control entailment 1:9}
\end{figure*}

\noindent\textbf{Additional Results under Different Model Access Settings.} 
Following Figure~\ref{fig: risk control mcqa Clopper–Pearson}, we also evaluate COIN-CP under the black-box setting on the CommonsenseQA dataset. 
As mentioned in Appendix~\ref{appendix: Details of Uncertainty Measures}, we sample 20 answers for each question and compute the normalized frequency score for each option, which is then employed to calculate the \textit{PE}. 
As shown in Figure~\ref{fig: commonsenseqa risk control cp black}, when model logits are inaccessible, COIN-CP can still rigorously constrain the ECFR on the test set, although several models' capability imposes limitations at lower risk levels. 
Moreover, we perform COIN on the TriviaQA dataset under the white-box setting. 
Figure~\ref{fig: triviaqa risk control cp white} suggests that, for some models, internal logits may be unreliable due to factors such as hallucination or overconfidence, resulting in larger ECFR variance compared to Figure~\ref{fig: risk control open-domain qa Clopper–Pearson}. 
Overall, regardless of whether the model's internal confidence information is accessible, COIN can rigorously control the selection risk. 

\noindent\textbf{Additional Results under Different Correctness Metrics.} 
We further evaluate the performance of COIN on the open-domain TriviaQA task, using bi-entailment to assess answer correctness. 
As demonstrated in Figures~\ref{fig: triviaqa risk control entailment} and \ref{fig: triviaqa risk control entailment 1:9}, both COIN-CP and COIN-HFD effectively control the risk. 
When the ratio of calibration data to test data is as low as 1:9, the ECFR variance increases; however, the mean consistently remains below the upper bound, highlighting COIN's predictive efficiency. 


\begin{figure*}[!t]
    \centering
    \includegraphics[width=\linewidth]{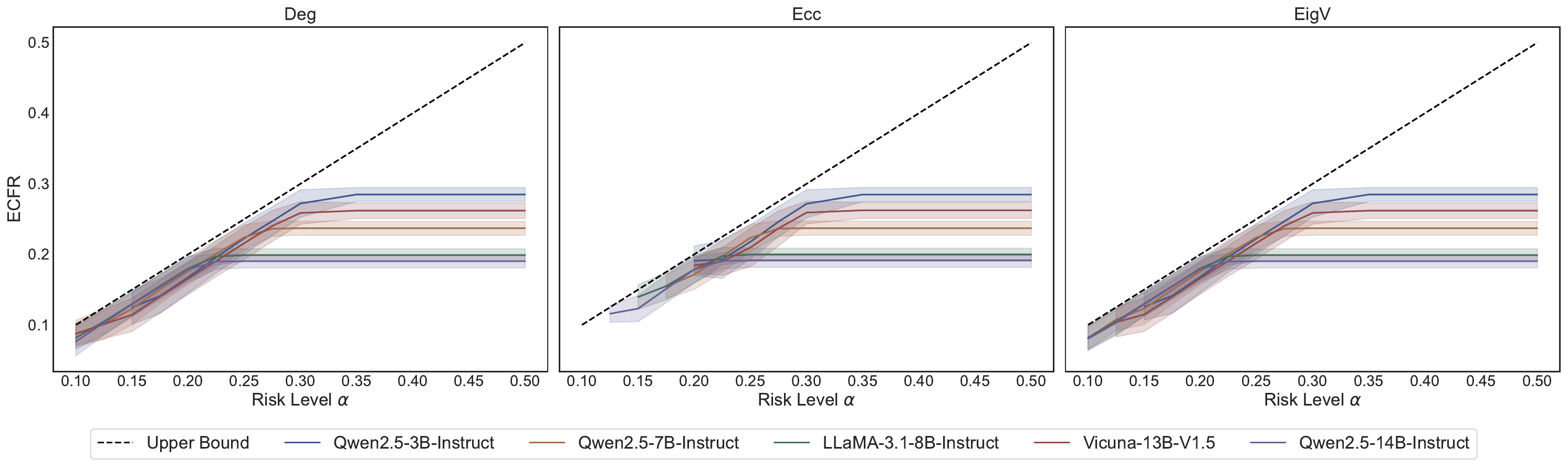}
    \caption{ECFR Results on TriviaQA, with sentence similarity as the correctness metric (1:1).}
    \label{fig: triviaqa risk control uncertainty combination similarity}
\end{figure*}

\begin{figure*}[!t]
    \centering
    \includegraphics[width=\linewidth]{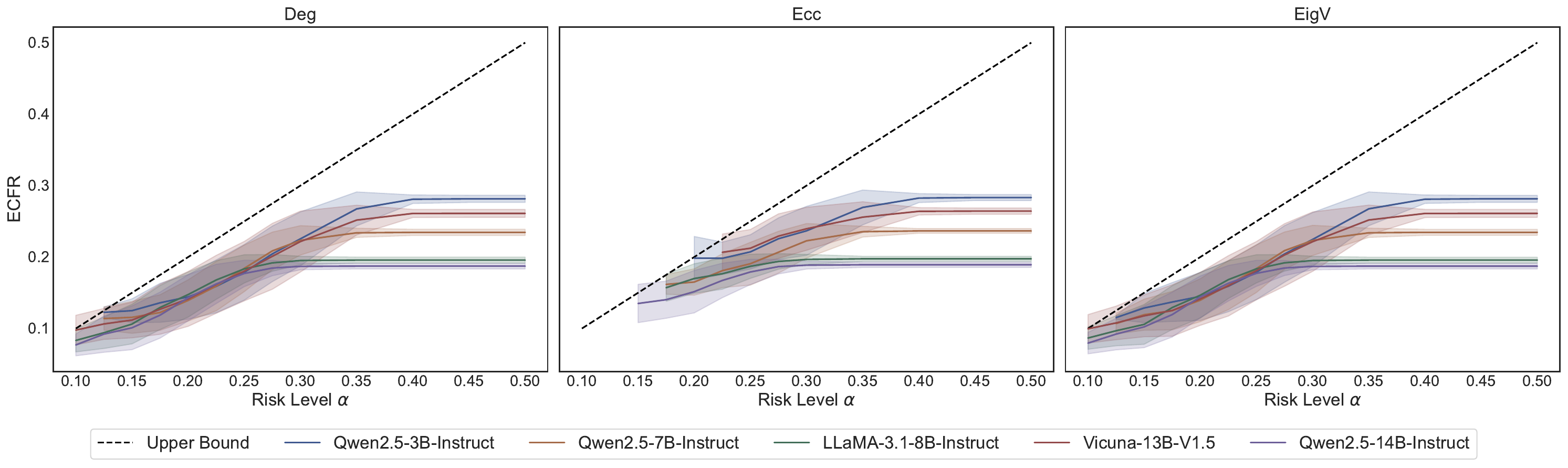}
    \caption{ECFR Results on TriviaQA, with sentence similarity as the correctness metric (1:9).}
    \label{fig: triviaqa risk control uncertainty combination similarity 1:9}
\end{figure*}

\begin{figure*}[!t]
    \centering
    \includegraphics[width=\linewidth]{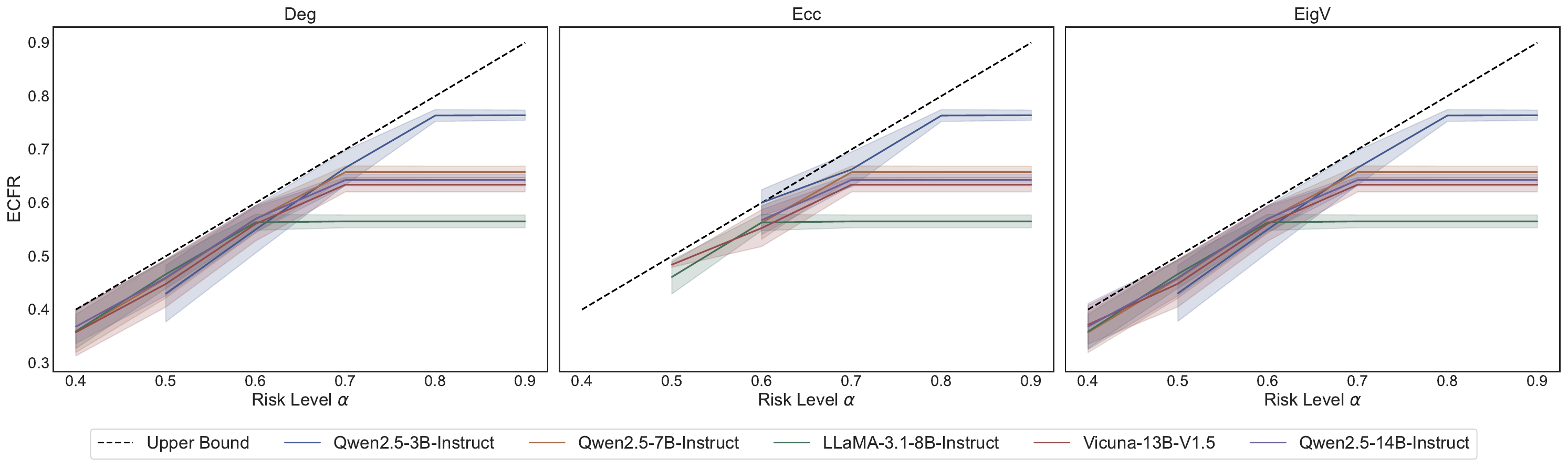}
    \caption{ECFR Results on TriviaQA, with bi-entailment as the correctness metric (1:1).}
    \label{fig: triviaqa risk control uncertainty combination entailment}
\end{figure*}

\begin{figure*}[!t]
    \centering
    \includegraphics[width=\linewidth]{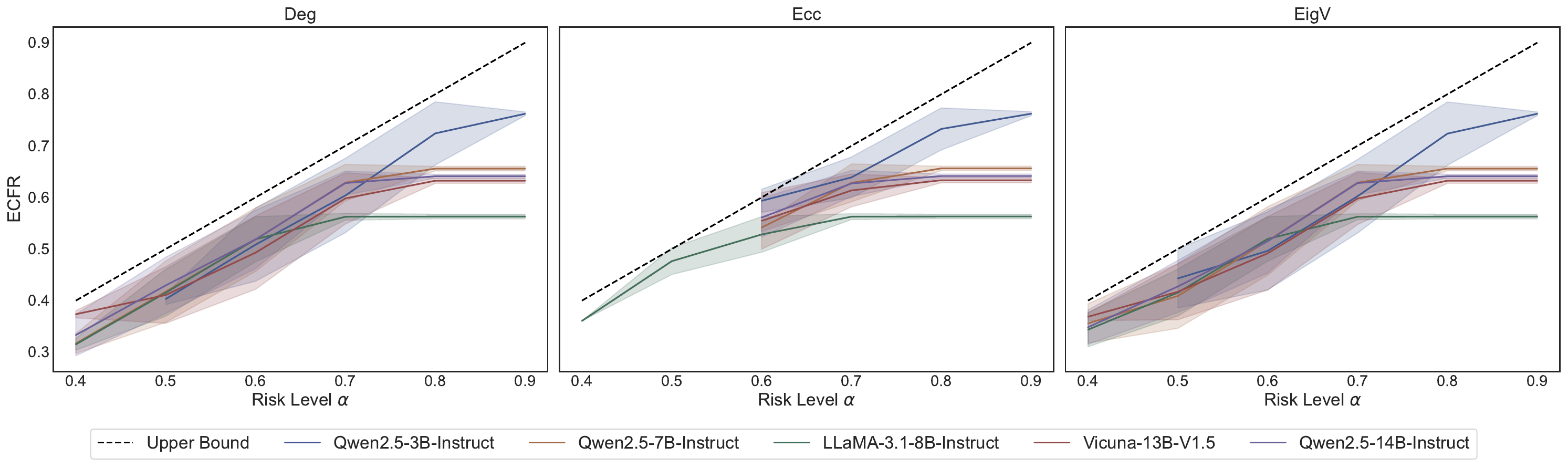}
    \caption{ECFR Results on TriviaQA, with bi-entailment as the correctness metric (1:9).}
    \label{fig: triviaqa risk control uncertainty combination entailment 1:9}
\end{figure*}

\begin{figure*}[t]
    \centering
    \begin{subfigure}{0.495\linewidth}
        \centering
        \includegraphics[width=\linewidth]{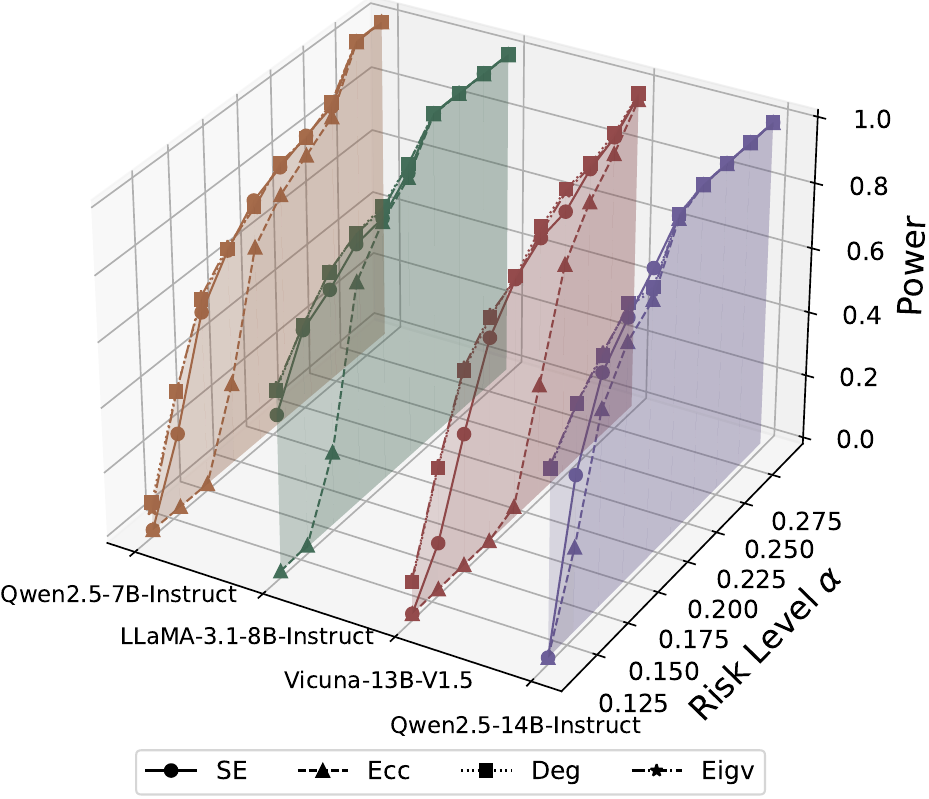}
        \caption{CommonsenseQA.}
	\label{fig: power triviaqa cp uncertainty}
    \end{subfigure}
    \hfill
    \centering
    \begin{subfigure}{0.495\linewidth}
	\centering
	\includegraphics[width=\linewidth]{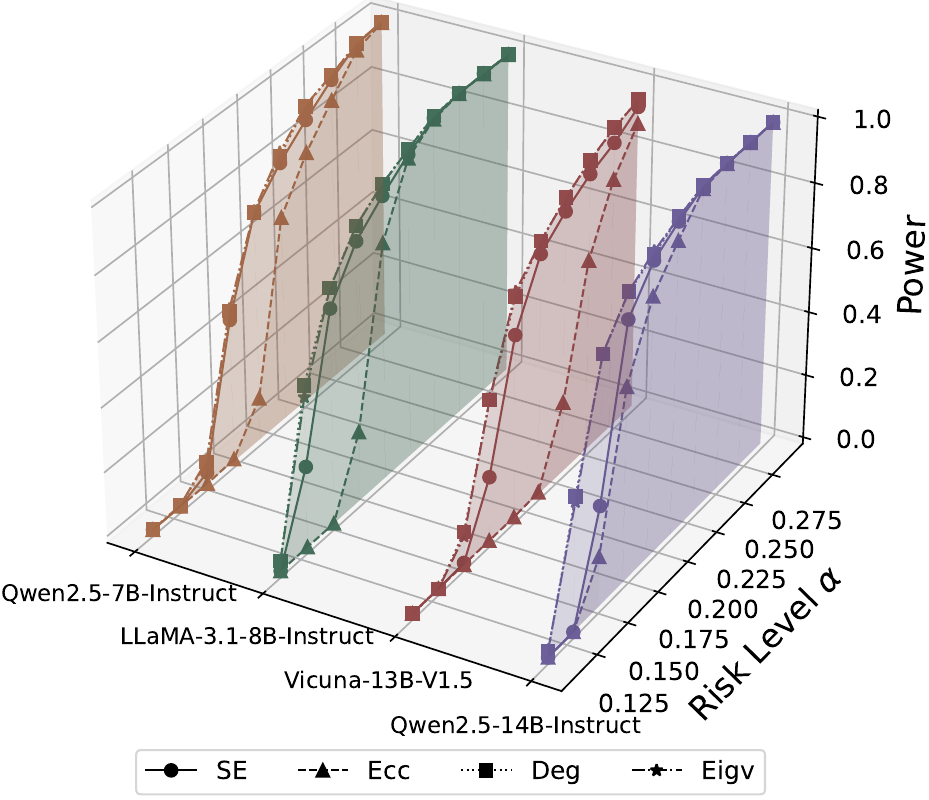}
	\caption{TriviaQA.}
	\label{fig: power triviaqa hfd uncertainty}
    \end{subfigure}
    \caption{Comparison of the power metric across various UQ approaches.}\label{fig: power triviaqa uncertainty comparison}
\end{figure*}

\noindent\textbf{Additional Results under Alternative UQ Strategies.} 
COIN is highly extensible, allowing more advanced UQ approaches to be incorporated in the first stage to enhance the accuracy of empirical error analysis, thereby improving power performance while maintaining risk control. 
Figures~\ref{fig: triviaqa risk control uncertainty combination similarity} to~\ref{fig: triviaqa risk control uncertainty combination entailment 1:9} illustrate that COIN-CP consistently maintains strict control over the ECFR metric on the TriviaQA dataset, across a wide range of LLMs, UQ methods, correctness evaluation metrics, and calibration-to-test data ratios. 
Additionally, Figure~\ref{fig: power triviaqa uncertainty comparison} demonstrates that leveraging more advanced UQ techniques in the first stage can significantly enhance the power of COIN to retain more correct answers on both closed-ended CommonsenseQA and open-domain TriviaQA datasets. 

\begin{figure*}[!h]
    \centering
    \includegraphics[width=0.6\linewidth]{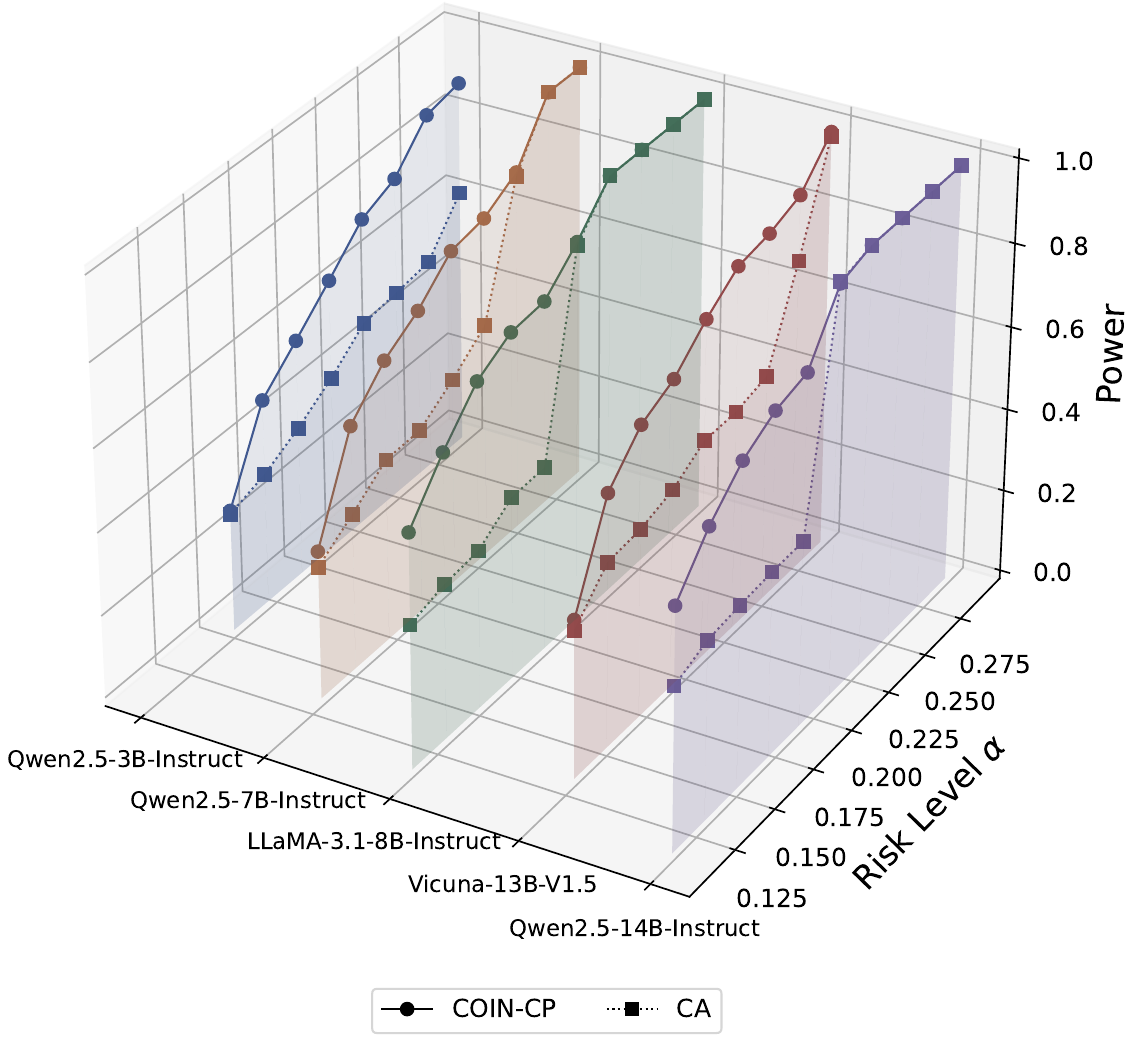}
    \caption{Comparison of the power performance on the TriviaQA dataset across five LLMs, utilizing Eigv as the UQ strategy.}
    \label{fig: power triviaqa eigv cp-ca}
\end{figure*}

\noindent\textbf{Power Comparison with CA when Using the EigV Approach.} 
As illustrated in Figure~\ref{fig: power triviaqa uncertainty comparison}, EigV achieves the best overall power performance across various user-specified risk levels. 
We further compare the power of COIN-CP and CA under this setting. 
As shown in Figure~\ref{fig: power triviaqa eigv cp-ca}, on the TriviaQA dataset, COIN-CP consistently outperforms CA across five LLMs. 
For example, at a low risk level of 0.11, COIN-CP achieves a power of 0.56 on the LLaMA-3.1-8B-Instruct model, surpassing CA by 0.21, significantly increasing the selection of correct answers under FDR constraint.

\begin{figure*}[!t]
    \centering
    \begin{subfigure}{0.495\linewidth}
        \centering
        \includegraphics[width=\linewidth]{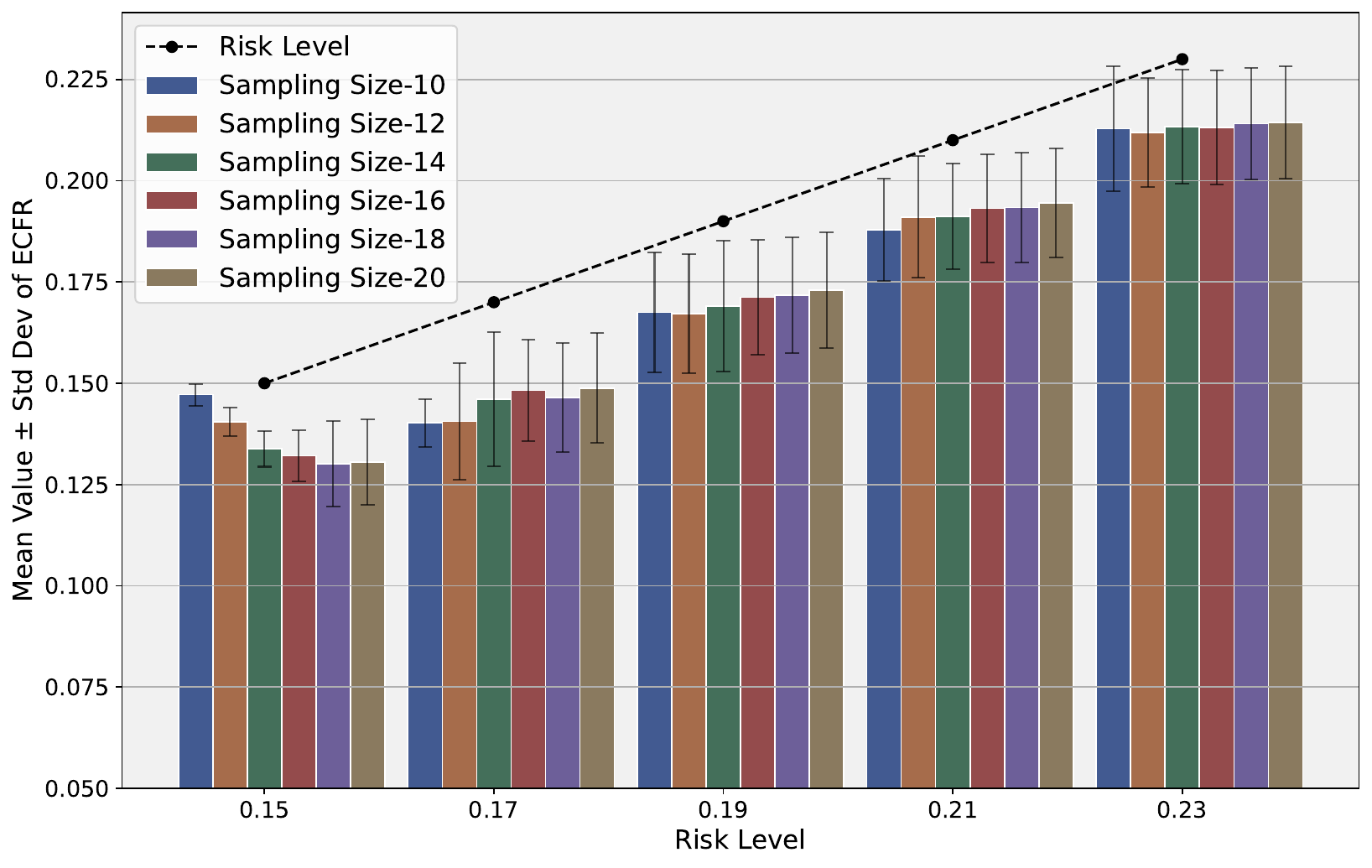}
        \caption{CommonsenseQA.}
	\label{fig: commonsenseqa cp sample}
    \end{subfigure}
    \hfill
    \centering
    \begin{subfigure}{0.495\linewidth}
	\centering
	\includegraphics[width=\linewidth]{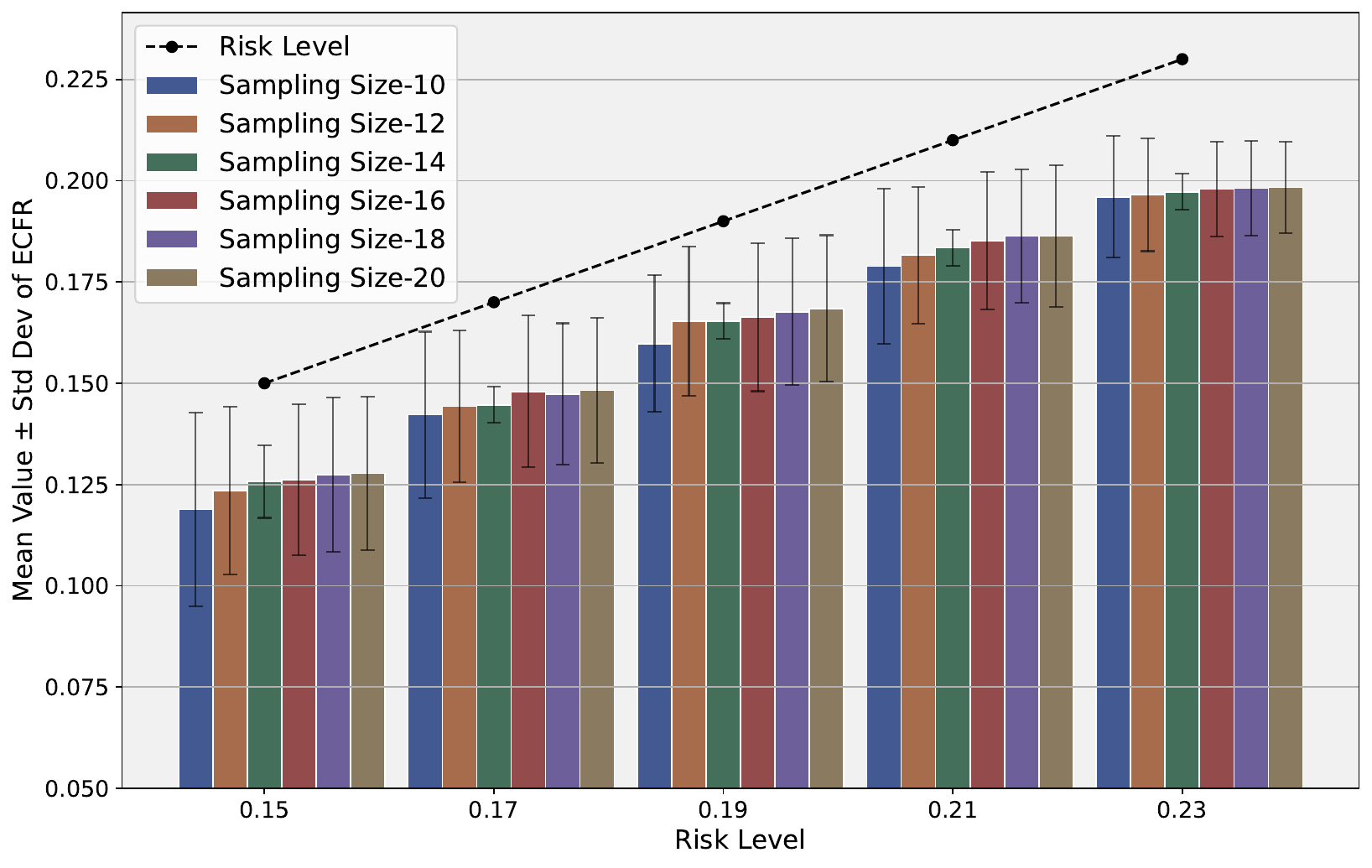}
	\caption{TriviaQA.}
	\label{fig: triviaqa cp sample}
    \end{subfigure}
    \caption{ECFR results across various sampling sizes for UQ in the first stage of COIN-CP.}\label{fig: sample cp ablation}
\end{figure*}

\begin{figure*}[!t]
    \centering
    \begin{subfigure}{0.495\linewidth}
        \centering
        \includegraphics[width=\linewidth]{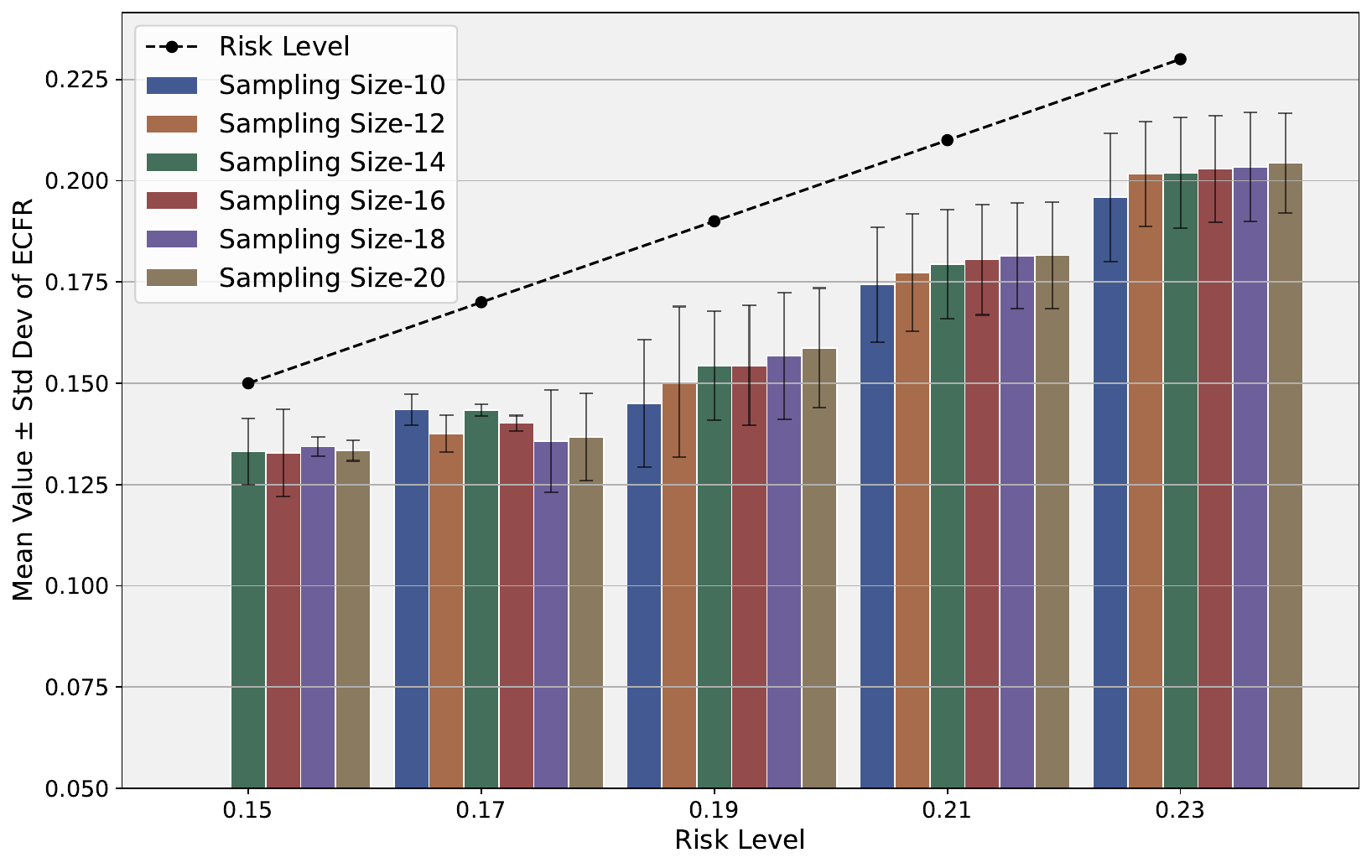}
        \caption{CommonsenseQA.}
	\label{fig: commonsenseqa hfd sample}
    \end{subfigure}
    \hfill
    \centering
    \begin{subfigure}{0.495\linewidth}
	\centering
	\includegraphics[width=\linewidth]{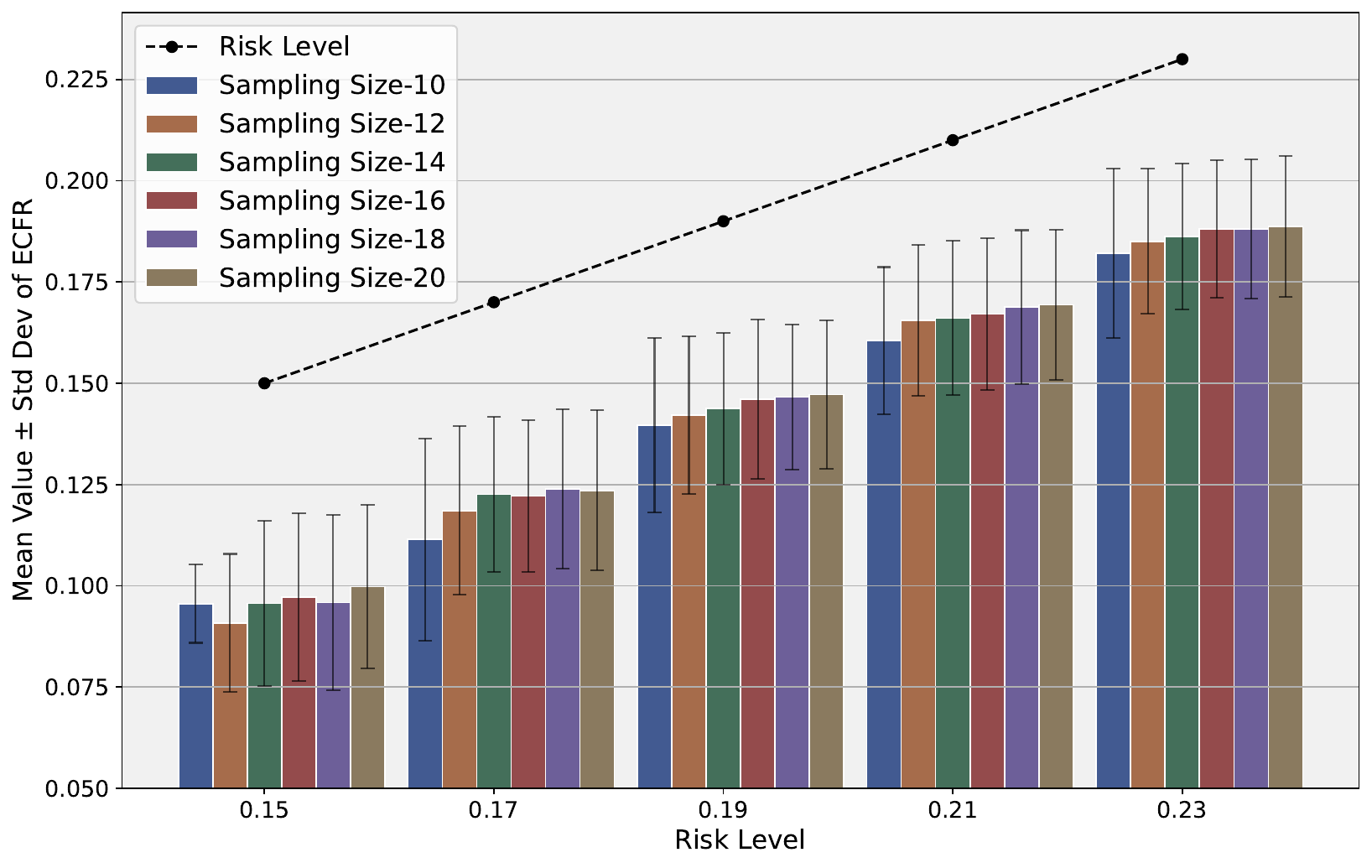}
	\caption{TriviaQA.}
	\label{fig: triviaqa hfd sample}
    \end{subfigure}
    \caption{ECFR results across various sampling sizes for UQ in the first stage of COIN-HFD.}\label{fig: sample hfd ablation}
\end{figure*}

\begin{figure*}[!t]
    \centering
    \begin{subfigure}{0.495\linewidth}
        \centering
        \includegraphics[width=\linewidth]{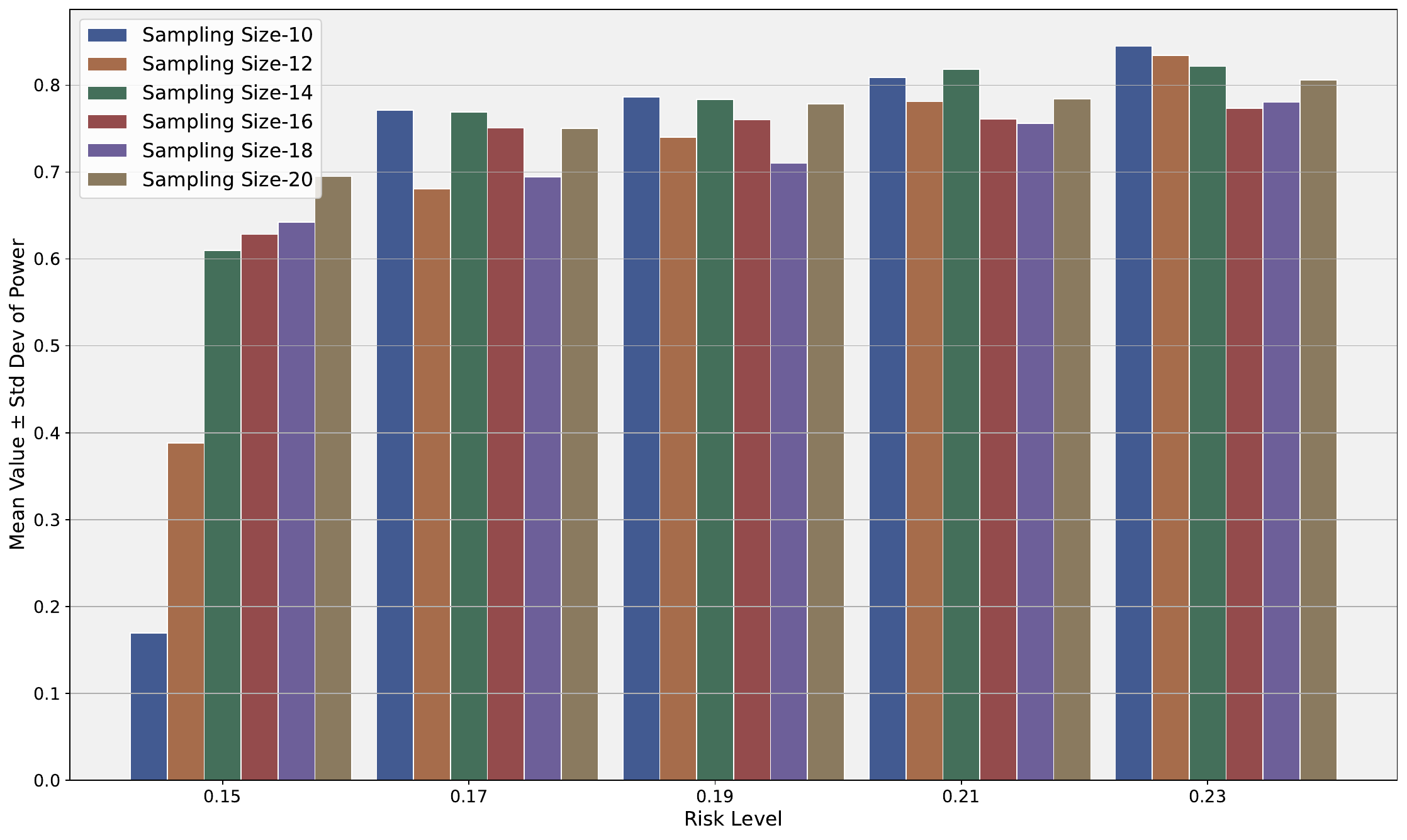}
        \caption{CommonsenseQA.}
	\label{fig: commonsenseqa cp power sample}
    \end{subfigure}
    \hfill
    \centering
    \begin{subfigure}{0.495\linewidth}
	\centering
	\includegraphics[width=\linewidth]{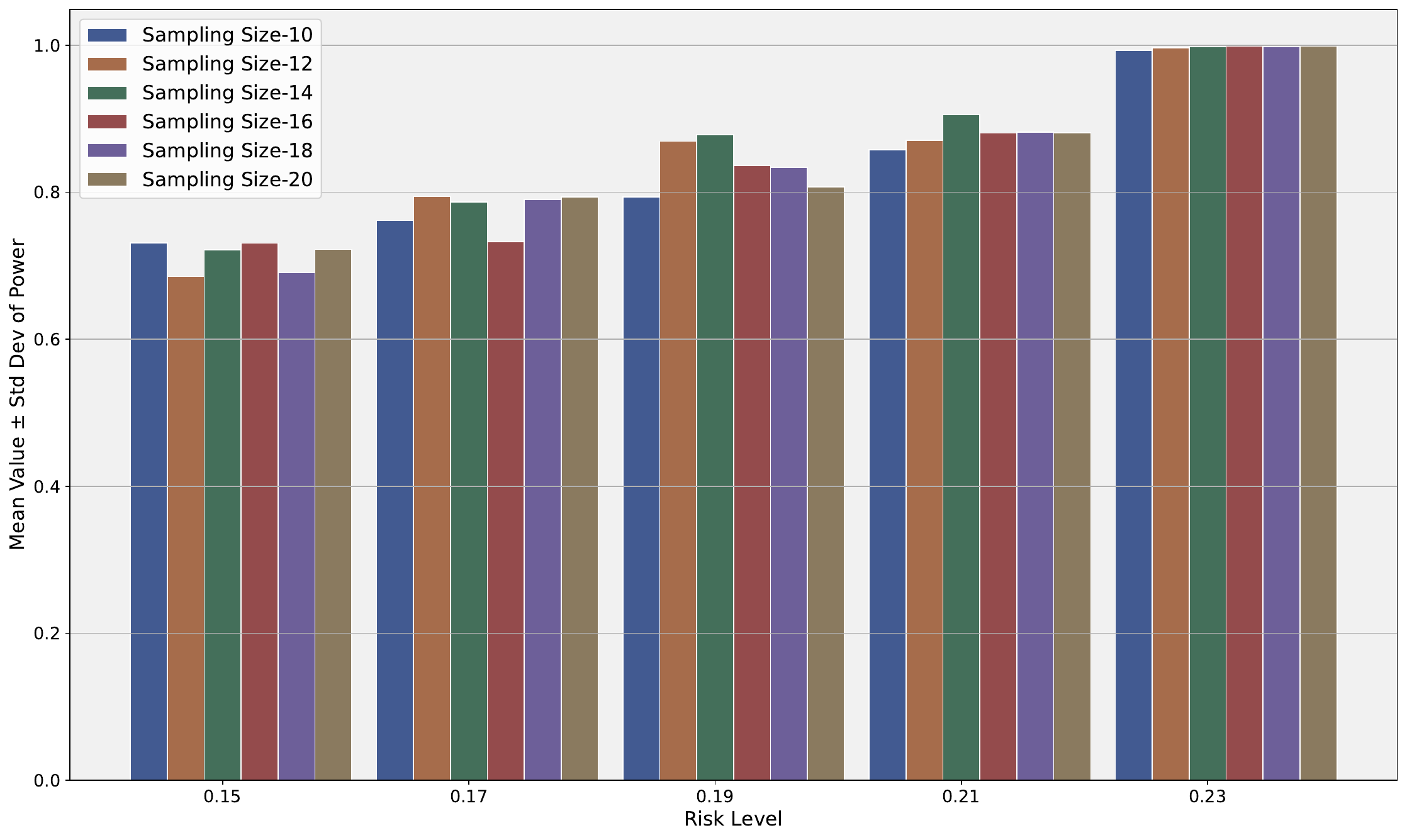}
	\caption{TriviaQA.}
	\label{fig: triviaqa cp power sample}
    \end{subfigure}
    \caption{Power results across various sampling sizes for UQ in the first stage of COIN-CP.}\label{fig: sample cp power ablation}
\end{figure*}

\noindent\textbf{Additional Evaluations across Various Sampling Sizes under Black-Box Settings.} 
In light of computational constraints in real-world QA scenarios, we investigate whether reducing the sampling size for the \textit{SE} method would compromise the risk control robustness of COIN. 
We utilize the LLaMA-3.1-8B-Instruct model. 
As demonstrated in Figure~\ref{fig: sample cp ablation} and Figure~\ref{fig: sample hfd ablation}, the sampling size is reduced from 20 to 10 on the CommonsenseQA dataset, and from 10 to 5 on the TriviaQA dataset, yet both COIN-CP and COIN-HFD consistently maintain control over the ECFR on test data across various desired risk levels. 
Besides risk control, we further examine whether reducing the sampling size would significantly degrade the power performance of COIN. 
As illustrated in Figure~\ref{fig: sample cp power ablation}, on the CommonsenseQA dataset, when the risk level is 0.15, reducing the sampling size from 20 to 14 does not lead to a significant drop in power performance. 
When the risk level exceeds 0.15, the power remains stable. 
On the TriviaQA dataset, the power performance remains unaffected by reductions in sampling size across all evaluated risk levels. 
These results underscore the practicality and efficiency of the COIN framework.

\newpage
\begin{figure*}[!t]
\begin{tcolorbox}[title=CommonsenseQA]
$\#\#\#$ System:\\
Make your best effort and select the correct answer for the following multiple-choice question. 
For each question, only one choice is correct. 
Answer should be one among A, B, C, D, E.
\\

$\#\#\#$ User:\\
What is something I need to avoid while playing ball?\\
A: competition\\
B: losing\\
C: injury\\
D: hitting the ball\\
E: having fun\\
$\#\#\#$ Assistant:\\
C\\

$\#\#\#$ User:\\
The sanctions against the school were a punishing blow, and they seemed to what the efforts the school had made to change?\\
A: ignore\\
B: enforce\\
C: authoritarian\\
D: yell at\\
E: avoid\\
$\#\#\#$ Assistant:\\
\end{tcolorbox}
\caption{A prompt example in the CommonsenseQA Dataset.}
\label{fig: CommonsenseQA prompt}
\end{figure*}

\begin{figure*}[!t]
\begin{tcolorbox}[title=TriviaQA]
$\#\#\#$ System:\\
This is a bot that correctly answers questions.
\\

$\#\#\#$ User:\\
In 1968, who did radical feminist Valerie Solanas shoot and wound as he entered his New York studio?\\
$\#\#\#$ Assistant:\\
Andy Warhol\\

$\#\#\#$ User:\\
Who was the man behind The Chipmunks?\\
$\#\#\#$ Assistant:\\
\end{tcolorbox}
\caption{A prompt example in the TriviaQA Dataset.}
\label{fig: TriviaQA prompt}
\end{figure*}

\begin{figure*}[!t]
\begin{tcolorbox}[title=MMVet]
$<$image$>$
\\
What is $x$ in the equation?
\\
NOTE: Provide only the final answer. Do not provide unrelated details.
\end{tcolorbox}
\caption{A prompt example in the MMVet Dataset.}
\label{fig: MMVet prompt}
\end{figure*}

\end{document}